\documentclass[10pt,journal]{IEEEtran}
\usepackage[T1]{fontenc}% optional T1 font encoding

\usepackage{bm}
\usepackage{amsmath}
\usepackage{float}
\usepackage{setspace}
\usepackage{amssymb}
\usepackage{stfloats}
\usepackage{cite}
\usepackage{ragged2e}
\usepackage[ruled,vlined,linesnumbered]{algorithm2e}
\usepackage{amsfonts}
\usepackage{mathrsfs}
\usepackage{amsthm}
\usepackage{array,booktabs}
\usepackage{multirow}
\usepackage{cuted}
\usepackage{multicol}
\usepackage{graphicx}
\usepackage{svg}
\usepackage{subfigure}
\usepackage{xcolor}
\usepackage{hyperref}
\usepackage{threeparttable}
\usepackage{dcolumn}
\usepackage{makecell}
\usepackage{lipsum}
\usepackage{enumerate}

\usepackage[table]{xcolor}

\graphicspath{{figures/}}
\svgpath{{figures/}}
\newcommand\MYhyperrefoptions{bookmarksnumbered=true,
	pdfpagemode={UseOutlines},plainpages=false,
	colorlinks=false,pdfborder={0 0 1},linkbordercolor={1 0 0},citebordercolor={0 1 0},urlbordercolor={0 1 1},
	pdftitle={SAGE: A Socially-Aware Generative Engine for Heterogeneous Multi-Agent Navigation},%
	pdfsubject={Heterogeneous multi-agent navigation},
	pdfauthor={Lan Hu, Minghui Liwang, Wenbo Zhu, Xinlei Yi, Yiguang Hong, Xianbin Wang, Zhenzhen Jiao, Seyyedali Hosseinalipour},
	pdfkeywords={socially-aware navigation, heterogeneous multi-agent systems, diffusion models, trajectory generation, safety-social guidance, heterogeneous graph transformer, multi-robot coordination}}
\expandafter\hypersetup\expandafter{\MYhyperrefoptions}

\usepackage{microtype}

\begin{document}
	\bstctlcite{IEEEtran:BSTcontrol}
	    
	\title{\vspace{-3mm} SAGE: A Socially-Aware Generative Engine for Heterogeneous Multi-Agent Navigation}
	
	\author{\vspace{-1mm} Lan Hu, Minghui Liwang, \textit{Senior Member, IEEE}, Wenbo Zhu, Xinlei Yi, \textit{Senior Member, IEEE}, Yiguang Hong, \textit{Fellow, IEEE}, Xianbin Wang, \textit{Fellow, IEEE}, Zhenzhen Jiao, Seyyedali Hosseinalipour, \IEEEmembership{Senior Member}, \IEEEmembership{IEEE}%
	\thanks{L. Hu (2453781@tongji.edu.cn) is with the Guohao School, Tongji University, Shanghai, China. M. Liwang (minghuiliwang@tongji.edu.cn), W. Zhu (wbzhu@tongji.edu.cn), X. Yi (xinleiyi@tongji.edu.cn) and Y. Hong (yghong@tongji.edu.cn) are with the Shanghai Research Institute for Intelligent Autonomous Systems, the State Key Laboratory of Autonomous Intelligent Unmanned Systems, Department of Control Science and Engineering, Tongji University, Shanghai, China. X. Wang (xianbin.wang@uwo.ca) is with the Department of Electrical and Computer Engineering, Western University, Ontario, Canada. Z. Jiao (jiaozhenzhen@gmail.com) is with Aeromind Technology Co., Ltd., Shenzhen, China. S. Hosseinalipour (alipour@buffalo.edu) is with the Department of Electrical Engineering, University at Buffalo-SUNY, USA. Corresponding author: M. Liwang.}
    \vspace{-5mm}}

	\IEEEtitleabstractindextext{
		\begin{abstract}
			\justifying
    Safe and socially compliant navigation in open human-robot environments requires robots to predict the motions of \textit{heterogeneous agents}, such as pedestrians, cyclists,  and other robots, while planning trajectories that satisfy physical safety requirements and social interaction norms.
    Nevertheless, existing navigation methods typically assume homogeneous interactions across agents/entities or enforce only geometric collision constraints, making them inadequate for modeling the interaction patterns among heterogeneous agents  and social-compliance constraints (i.e., maintaining appropriate interpersonal distances). Motivated by this, we propose SAGE, a \underline{s}ocially-\underline{a}ware \underline{g}enerative \underline{e}ngine for heterogeneous multi-agent navigation.  SAGE represents robots and surrounding agents as a \textit{directed heterogeneous graph} and employs a \textit{heterogeneous graph transformer} (HGT) to learn interaction representations that distinguish different agent types and interaction patterns. Conditioned on this representation, we develop a \textit{diffusion-based generative model} that jointly generates surrounding-agent trajectory predictions and robot trajectory plans within a unified probabilistic framework. During inference, we introduce a \textit{training-free safety-social energy guidance mechanism} that injects differentiable collision-avoidance, kinematic-feasibility, task-progress, and role-dependent social-compliance objectives into the diffusion sampling process, thereby refining the generated robot trajectories without retraining. Through experiments on real-world (ETH/UCY and SDD) and synthetic datasets, we demonstrate that SAGE consistently improves safety and social compliance while maintaining task performance; further, its training-free guidance mechanism reduces collision and social-violation rates, scales to teams of up to 20 robots, and enables control of the trade-off among safety, task performance, and trajectory accuracy. 
		\end{abstract}

        \vspace{-1mm}
}
	
	%}
\maketitle
\IEEEdisplaynontitleabstractindextext
% \fi
%\IEEEpeerreviewmaketitle
\IEEEpeerreviewmaketitle

\vspace{-3mm}
\section{Introduction}

\IEEEPARstart{D}{riven} by recent advances in large-scale robotic systems~\cite{karoly2021deep}, autonomous agents are rapidly transitioning from structured and controlled environments to unstructured, open environments characterized by dense human-robot coexistence and interaction. In such mixed-traffic scenarios, achieving safe, efficient, and socially compliant autonomous navigation remains a major challenge~\cite{kretzschmar2016socially,chen2022interactive}. This is because unlike structured environments, these scenarios feature heterogeneous agents with diverse behavioral patterns and interactions. Specifically, these agents can be classified into four categories based on their levels of autonomy and control authority: \textit{(i) Purely human agents} (PHAs, e.g., pedestrians and runners), who are driven entirely by human will, exhibit highly random behavior and adhere to a wide range of social interaction norms~\cite{helbing1995social}; \textit{(ii) Human-controlled non-autonomous agents} (HNAAs, e.g., skateboarders and cyclists), who rely on manual power or simple mechanical propulsion without autonomous sensing, making their trajectories highly dependent on users' real-time reflexes; \textit{(iii) Shared-control semi-autonomous agents} (SSAAs, e.g., human-driven vehicles with advanced driver-assistance systems), where human decision-making operates in tandem with intelligent assistance, retaining inherent behavioral unpredictability despite partial rule compliance~\cite{schwarting2019social}; and \textit{(iv) Execution-autonomous agents} (EAAs, e.g., automated robots), which are data-driven, with their behavior being relatively predictable under limited compatibility with human social norms~\cite{Mavrogiannis2021CoreCO}. Among these four categories, PHAs, HNAAs, and SSAAs are independently behaving entities whose future motions cannot be directly controlled, requiring robots to continuously anticipate their future behaviors and interactions~\cite{gupta2018social}. In contrast, EAAs correspond to the controllable robots whose trajectories must be planned according to these predicted behaviors while simultaneously satisfying physical safety and social interaction requirements.

This tight coupling between \textit{environment prediction} and \textit{robot trajectory planning} is a defining characteristic of socially-aware navigation in heterogeneous environments. However, jointly solving these two tasks (i.e., predicting the future behaviors of surrounding heterogeneous agents and planning safe and socially compliant trajectories for autonomous robots) remains  challenging because robots must reason about human-driven and data-driven agents that exhibit diverse motion dynamics, autonomy levels, and social interaction behaviors. Addressing these diverse interactions requires navigation frameworks capable of reasoning about both heterogeneous agent behaviors and their influence on robot decision-making. Nevertheless, existing trajectory prediction and planning methods typically rely on homogeneous interaction patterns~\cite{alahi2016social,mohamed2020social} and/or solely enforce geometric collision avoidance~\cite{van2011reciprocal,zhou2017collision}, making them incapable of capturing the heterogeneous agent interactions and the associated social behaviors. Consequently, achieving safe and socially compliant navigation in heterogeneous multi-agent environments remains an open research area~\cite{rudenko2020human,Mavrogiannis2021CoreCO,li2024game,yuan2021agentformer}.

\vspace{-3mm}

\subsection{Core Motivation}
To address the aforementioned challenges, we investigate the following key research questions (RQs):

\noindent
$\bullet$ \textit{RQ 1: How can we develop principled models that capture structural and behavioral asymmetries in heterogeneous multi-agent interactions, where agents differ in dynamics, perception, and influence over the shared environment?}
Most mainstream trajectory generation models are developed under the homogeneous-agent assumption, employing shared parameters to uniformly model diverse agent types~\cite{alahi2016social,mohamed2020social}. However, in real-world environments, interactions vary significantly across agent categories. For example, a robot may adopt proactive yielding when encountering
pedestrians, yet switch to predictive trajectory tracking when
navigating around vehicles. Ignoring these \textit{role-dependent} interactions limits the ability of existing models to accurately predict agent trajectories and plan robot trajectories. Since these interaction asymmetries naturally arise from the semantic roles of different agents and their relationships, they can be represented as a \textit{heterogeneous interaction graph}, where nodes correspond to different agent categories and edges encode their interaction relationships~\cite{kosaraju2019social}. Based on this  representation, we develop a \textit{heterogeneous graph transformer} (HGT), inspired by~\cite{hu2020heterogeneous}, that learns distinct interaction patterns across different agent categories. By explicitly distinguishing semantic types of nodes and edges, HGT learns independent attention projection matrices for each interaction relationship (e.g., ``robot$\mathord{\rightarrow}$pedestrian'', ``pedestrian$\mathord{\rightarrow}$robot'', and ``robot$\mathord{\rightarrow}$robot''), thereby capturing asymmetric interactions and producing socially-aware scene representations.

\noindent
$\bullet$ \textit{RQ 2: How can physical feasibility constraints and culturally induced behavioral norms be jointly modeled within a unified framework that reconciles hard kinematic limitations with context-dependent social regularities?}
Most existing generative models primarily enforce hard physical constraints, such as collision avoidance and kinematic feasibility, while largely overlooking soft social constraints. Nevertheless, in human-robot coexistence scenarios, navigation should ensure physical safety, while complying with social norms and human comfort expectations~\cite{hall1966hidden}. For instance, even if a trajectory is collision-free, it may still be socially unacceptable when it frequently intrudes into pedestrians' personal space or disrupts ongoing social interactions~\cite{hall1966hidden}. To address this, we employ a \textit{conditional diffusion model}~\cite{ho2020denoising,gu2022stochastic} as the generative backbone and propose a \textit{safety-social energy guidance mechanism} based on a \textit{differentiable energy function}~\cite{song2020score} during the inference stage. Specifically, social norms, such as maintaining comfortable interpersonal distances and exhibiting polite avoidance behaviors, are formulated as \textit{heterogeneous anisotropic potential fields}, whose gradients are incorporated into each diffusion denoising step to iteratively steer the generated trajectories toward lower-energy solutions. Unlike methods based on conditional variational autoencoders (CVAEs) and generative adversarial networks (GANs)~\cite{salzmann2020trajectron++,gupta2018social}, where constraints are implicitly learned during training and are difficult to modify afterward, diffusion models enable differentiable guidance to be injected directly into the sampling process. Consequently, new physical or social constraints can be incorporated or removed at inference time without retraining the generative model. Through this safety-social energy guidance, generated trajectories are encouraged toward kinematic feasibility~\cite{campion1996structural} and social compliance in a training-free  manner.

\vspace{-4mm}
\subsection{Novelty and Contribution}
To address the aforementioned research questions, we develop a \textit{\underline{s}ocially-\underline{a}ware \underline{g}enerative \underline{e}ngine}, called SAGE, that integrates heterogeneous interaction modeling, diffusion-based trajectory generation, and safety-social guidance. The key contributions of SAGE are summarized as follows:

\noindent
$\bullet$ \textit{A novel problem formulation for safe and socially compliant navigation in heterogeneous multi-agent environments.}
We formulate a new navigation problem in which autonomous robots operate alongside heterogeneous entities (e.g., pedestrians, cyclists, and semi-autonomous vehicles) exhibiting distinct dynamics, autonomy levels, and social behaviors. Unlike existing formulations, our problem explicitly captures the tight coupling among heterogeneous agent interactions, surrounding agent trajectory prediction, robot trajectory planning, and the joint consideration of hard physical safety constraints and soft social interaction norms. This unified formulation provides the foundation for developing socially-aware navigation algorithms in heterogeneous environments.

\noindent
$\bullet$ \textit{A unified generative framework for heterogeneous interaction modeling, trajectory prediction, and robot trajectory planning.}
We develop SAGE, a socially-aware generative framework that unifies heterogeneous interaction modeling, surrounding agent trajectory prediction, and robot trajectory planning within a single probabilistic framework. Specifically, SAGE represents heterogeneous agents and their interactions using a directed heterogeneous graph and develops an HGT to explicitly learn role-dependent and asymmetric interaction patterns across different agent categories. Building upon the resulting socially-aware scene representations, SAGE further introduces a diffusion-based generative model that simultaneously predicts the future trajectories of surrounding agents and generates robot trajectory plans, thereby explicitly capturing the coupling between environment prediction and robot motion planning.

\noindent
$\bullet$ \textit{A training-free safety-social guidance framework for controllable robot trajectory generation.}
We integrate a novel safety-social energy guidance mechanism in SAGE that enables explicit constraint-aware correction during diffusion-based trajectory generation. Specifically, we formulate collision avoidance, kinematic feasibility, task progress, and role-dependent social norms as differentiable energy functions, where heterogeneous anisotropic potential fields capture category-specific social behaviors. By injecting the resulting energy gradients into each diffusion denoising step, the proposed guidance iteratively reduces physical-safety and social-compliance violations at inference time, enabling controllable correction without modifying or retraining the underlying generative model.

\noindent
$\bullet$ \textit{Empirical validation.}
We validate the effectiveness of SAGE through experiments on real-world (ETH/UCY and SDD) and synthetic datasets, demonstrating its notable performance on reducing personal space intrusion and collision rates while maintaining trajectory diversity and kinematic feasibility.

\vspace{-1mm}
\section{Literature Review} 
In the following, we review the related literature while highlighting differences between prior studies and this work.

\vspace{-2mm}
\subsection{Trajectory Prediction for Socially-Aware Navigation}
Safe robot navigation in mixed human-robot environments relies on accurate prediction of surrounding entities' future trajectories for collision avoidance and motion planning. Early approaches predominantly used recurrent neural networks (RNNs), with Social-LSTM~\cite{alahi2016social} introducing social pooling (i.e., aggregating hidden states across neighboring agents) to capture inter-agent interactions and Social-GAN~\cite{gupta2018social} incorporating adversarial learning to model multimodal human motion. Subsequent works advanced these early studies by adopting attention and graph-based architectures to capture richer interactions. For instance, SoPhie~\cite{sadeghian2019sophie} and Social-BiGAT~\cite{kosaraju2019social} combined physical scene context with social attention, while Social-STGCNN~\cite{mohamed2020social} modeled interactions through spatio-temporal graphs. Trajectron++~\cite{salzmann2020trajectron++} introduced dynamic graphs to accommodate varying agent populations, and AgentFormer~\cite{yuan2021agentformer} employed self-attention to jointly capture agent-agent and agent-environment dependencies. Despite these advances, existing methods generally assume similar interaction mechanisms across agent categories or use lightweight semantic embeddings that cannot fully capture distinct behavioral and social interaction patterns. Moreover, they focus on trajectory prediction rather than jointly predicting surrounding-agent trajectories and planning robot motion, limiting their applicability to socially-aware navigation in heterogeneous environments.

% Motivated by these shortcomings, a body of work adopted attention mechanisms and graph-based architectures to capture richer interaction structures. For instance, SoPhie~\cite{sadeghian2019sophie} and Social-BiGAT~\cite{kosaraju2019social} integrated physical scene contexts with social attention for expressive prediction. Social-STGCNN~\cite{mohamed2020social} reformulated interactions as spatio-temporal graphs, achieving significant gains in inference efficiency. Trajectron++~\cite{salzmann2020trajectron++} introduced dynamic graph structures to handle variable agent counts. AgentFormer~\cite{yuan2021agentformer} used self-attention to jointly model agent-agent and agent-environment dependencies. Despite their improved prediction accuracy, existing methods generally assume similar interaction mechanisms across different agent categories or rely on lightweight semantic embeddings that cannot fully characterize their distinct behavioral and social interaction patterns. In addition, most of these approaches are designed solely for trajectory prediction, rather than jointly reasoning about surrounding-agent trajectory prediction and robot trajectory planning, limiting their applicability to socially-aware robot navigation in heterogeneous environments.

These limitations motivate \textit{heterogeneous graph learning}, which explicitly distinguishes agent categories and their interaction relationships. For example,  heterogeneous graph neural networks (GNNs)~\cite{zhang2019heterogeneous} and HGT~\cite{hu2020heterogeneous} learn type-specific representations for different nodes and edges, yet their application to socially-aware robot navigation remains largely unexplored. While recent works have considered task-motion planning and heterogeneous multi-robot coordination in dynamic environments~\cite{miloradovic2022gmp,faroni2024optimal}, heterogeneous graph learning has rarely been used to jointly reason about surrounding-agent trajectory prediction and robot trajectory planning. To bridge this gap, we develop an HGT-based framework that learns semantic, social, and kinematic representations of heterogeneous agents, providing a socially-aware scene representation that supports both surrounding-agent trajectory prediction and downstream robot trajectory planning.

% The above limitations suggest that heterogeneous graph learning can provide a promising direction for modeling multi-agent interactions, as it can explicitly distinguish different agent categories and their interaction relationships. For example, Frameworks such as heterogeneous graph neural networks (GNNs)~\cite{zhang2019heterogeneous} and the HGT~\cite{hu2020heterogeneous} provide powerful representation learning capabilities by assigning distinct representations to different node and edge types. However, their application to socially-aware robot navigation remains largely unexplored.  Specifically, while recent studies have highlighted the importance of jointly considering task-motion planning and heterogeneous multi-robot coordination in dynamic environments~\cite{miloradovic2022gmp,faroni2024optimal}, heterogeneous graph learning has rarely been leveraged to jointly reason about surrounding-agent trajectory prediction and robot trajectory planning. To bridge this gap, we develop an HGT-based interaction modeling framework that learns rich semantic, social, and kinematic representations of heterogeneous agents, providing a socially-aware scene representation that supports both surrounding-agent trajectory prediction and downstream robot trajectory planning.

\vspace{-3.2mm}
\subsection{Diffusion-Based Trajectory Generation for Robot Navigation}
\vspace{-.4mm}

Diffusion probabilistic models~\cite{sohl2015deep,ho2020denoising} have emerged as a powerful generative paradigm for flexible and controllable trajectory synthesis through iterative denoising. In robotics, Diffuser~\cite{janner2022planning} and Decision Diffuser~\cite{ajay2022conditional} formulated trajectory planning as conditional diffusion for long-horizon decision making. For trajectory generation, MID~\cite{gu2022stochastic} modeled human-motion uncertainty, while MotionDiffuser~\cite{jiang2023motiondiffuser} extended diffusion to controllable multi-agent prediction. Subsequent works incorporated scene constraints for collision-free generation~\cite{xiao2023safediffuser}, controllable traffic simulation~\cite{zhong2023guided}, and interactive robot navigation~\cite{niedoba2023diffusion}.
Recently, joint prediction--planning diffusion (JPPD)~\cite{wu2026jppd} jointly sampled single-robot and surrounding-participant trajectories with an occupancy-based differentiable safety potential during conditional flow-matching inference. However, it does not explicitly consider multiple controllable robots, robot-entity and robot-robot interactions, or role-conditioned social distances. 
Collectively, these studies establish diffusion models as a flexible framework for controllable trajectory generation, which remains an active topic in cybernetic systems~\cite{lin2026galc}.
Advancing this literature, we develop a differentiable safety-social energy guidance framework that injects collision avoidance, kinematic feasibility, task progress, and role-dependent social norms directly into diffusion sampling. By formulating these objectives as differentiable energy functions, their gradients iteratively guide denoising toward physically feasible and socially compliant robot trajectories; moreover, additional physical or social objectives can be incorporated as new energy terms at inference time without retraining the underlying diffusion model.

\vspace{-2mm}

\vspace{-1.8mm}
\subsection{Social Norm Modeling and Human-Aware Robot Navigation}
\vspace{-.4mm}

Socially-aware navigation enables robots to navigate while respecting implicit human social norms, such as maintaining appropriate interpersonal distances~\cite{kruse2013human,Mavrogiannis2021CoreCO}. Early approaches relied on hand-crafted models: the Social Force Model~\cite{helbing1995social} modeled interactions through attractive and repulsive forces, while ORCA~\cite{van2011reciprocal} used reciprocal velocity optimization for efficient collision avoidance, often resulting in conservative and mechanically rigid behaviors. More recent data-driven approaches include deep reinforcement learning (DRL)-based methods such as SARL~\cite{chen2019crowd} and interactive model predictive control (MPC) for dense-crowd navigation~\cite{chen2022interactive}. Recent studies have further incorporated richer social cues, including group behavior and proxemics~\cite{hall1966hidden}, while Transformer-based methods have advanced promptable human trajectory prediction~\cite{saadatnejad2023social}. Concurrently, evaluation frameworks increasingly assess socially-aware navigation beyond trajectory accuracy by considering safety, comfort, and social compliance~\cite{gao2022evaluation,biswas2022socnavbench,saadatnejad2021socially}.

% Despite these advances, existing socially-aware navigation methods primarily focus on learning socially plausible behaviors through policy optimization or trajectory prediction, rather than explicitly incorporating social principles into the trajectory generation process. Consequently, social norms are often implicitly encoded in learned model parameters, making them difficult to interpret, modify, or adapt to different environments and interaction contexts. Moreover, most existing approaches emphasize surrounding-agent trajectory prediction, with limited integration of downstream robot trajectory planning within a unified generative framework. To address these limitations, we formulate physical safety objectives and social interaction principles, including interpersonal comfort distances and category-dependent social margins, as differentiable energy functions that directly guide diffusion-based trajectory generation. By injecting their gradients into the diffusion sampling process, SAGE favors physical feasibility and social compliance in a training-free and controllable manner, avoiding both opaque black-box supervision and rigid rule-based engineering.

Despite these advances, the above methods primarily learn socially plausible behaviors through policy optimization or trajectory prediction, leaving social norms implicitly encoded in model parameters and difficult to interpret, modify, or adapt. Moreover, most approaches focus on surrounding-agent trajectory prediction with limited integration of robot trajectory planning within a unified generative framework. To address these limitations, we formulate physical safety and social interaction principles, including interpersonal comfort distances and category-dependent social margins, as differentiable energy functions that directly guide diffusion-based trajectory generation. By injecting their gradients during diffusion sampling, SAGE promotes physical feasibility and social compliance in a training-free and controllable manner, avoiding opaque black-box supervision and rigid rule-based engineering.

% Beyond human-robot interactions, multi-robot coordination has also been investigated, including switching-topology coordination~\cite{wen2021coordination}, distributed collision and deadlock avoidance~\cite{zhou2017collision}, swarm formation control~\cite{wu2022distributed}, time-constrained task allocation~\cite{turner2018distributed}, and learning-based flocking with repulsive interactions~\cite{bai2024learning}. While these studies provide the foundation for coordinated robot behavior, they typically treat robot-robot and human-robot coordination separately. In contrast, SAGE unifies these two perspectives by representing robot-robot and robot-entity interactions within the same heterogeneous graph and regulating them through a common safety-social energy guidance mechanism, enabling coordinated, socially-aware navigation for heterogeneous multi-agent systems.

Beyond human-robot interactions, multi-robot coordination has been studied through switching-topology coordination~\cite{wen2021coordination}, distributed collision and deadlock avoidance~\cite{zhou2017collision}, swarm formation~\cite{wu2022distributed}, task allocation~\cite{turner2018distributed}, and learning-based flocking~\cite{bai2024learning}. However, these studies typically treat robot-robot and human-robot coordination separately. 
In contrast, SAGE models both robot-robot and robot-entity interactions within a unified heterogeneous graph and regulates them through a safety-social energy guidance mechanism, enabling socially-aware navigation in heterogeneous multi-agent systems.

% In contrast, SAGE models both robot-robot and robot-entity interactions within a unified heterogeneous graph and regulates them through a common safety-social energy guidance mechanism, enabling coordinated and socially-aware navigation in heterogeneous multi-agent systems.

\vspace{-2mm}
\section{Core Models and Problem Formulation}
% \vspace{}

In heterogeneous urban environments characterized by evolving human-machine interactions, robot navigation necessitates the tight coupling of heterogeneous-entity trajectory prediction and socially-aware robot trajectory planning. We formalize this problem under dynamic multi-agent interactions, where each robot, i.e., EAA, is assigned a navigation task such as moving from its current/start state toward an assigned waypoint while satisfying both physical feasibility and social compliance. Meanwhile, surrounding heterogeneous entities, including PHAs, HNAAs, and SSAAs, exhibit  evolving and uncontrollable behaviors, requiring the robot to continuously anticipate their future motions and interaction patterns for reliable navigation\footnote{In this work, we use \textit{navigation} to denote the overall task, \textit{prediction} to describe the future trajectories of entities, \textit{trajectory planning} to describe the trajectory-planning problem for the main robots, and \textit{trajectory generation} to describe the diffusion-based solution process.}. For clarity, we use \textit{agents} as a general term for all participants, \textit{robots} to denote controllable agents, \textit{entities} to describe heterogeneous non-robot participants, and \textit{neighbors} to represent entities within the perception neighborhood of a specific robot. As illustrated in Fig.~\ref{fig:overview}, SAGE, which is a multi-agent social perception and navigation framework, consists of the following three synergistic phases:

\begin{figure}[!t]
\vspace{-3.5mm}
    \centering
    \includegraphics[width=.8\linewidth]{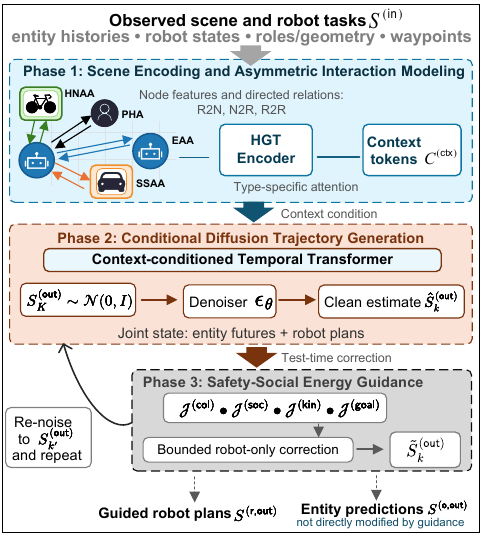}
    \vspace{-4.25mm}
    \caption{Overview of the three phases of SAGE. Phase~1 encodes scene histories, agent attributes, and robot waypoints into context tokens using a directed heterogeneous graph and HGT. At each reverse diffusion step, Phase~2 produces a clean estimate of the joint robot--entity trajectories, and Phase~3 applies bounded robot-only energy guidance. The corrected estimate is then re-noised for the next reverse step, and this cycle repeats until yielding the final guided robot plans and entity trajectory predictions.}
    \label{fig:overview}
    \vspace{-7.5mm}
\end{figure}

\noindent
$\bullet$ \textit{Phase 1: Heterogeneous scene encoding and interaction modeling.} 
Unlike traditional trajectory planning methods that treat agents homogeneously, this phase integrates neighbors' historical trajectories with semantic and kinematic attributes (such as pedestrian vulnerability and vehicle motion characteristics). Through a \textit{directed heterogeneous interaction graph}, the framework captures asymmetric interaction dynamics and implicit social interactions, enabling socially-aware navigation behaviors, including pedestrian yielding and predictive vehicle interaction. The resulting scene representation, which utilizes HGT, provides structured social semantics for downstream robot trajectory generation.

% \noindent
% $\bullet$ \textit{Phase 2: Joint trajectory generation via conditional diffusion.} 
% Given the inherent uncertainty in future entity behaviors, this phase does not produce a single deterministic trajectory. Instead, we leverage a conditional diffusion model to generate a multimodal joint trajectory distribution that simultaneously encompasses both the predicted future behaviors of heterogeneous entities and the preliminary task-directed trajectory plans of robots. This establishes a tightly coupled prediction-planning paradigm, capturing robot-environment interaction dynamics at a distribution-level. 

\noindent
$\bullet$ \textit{Phase 2: Joint trajectory generation via conditional diffusion.} 
Given the inherent uncertainty in future entity behaviors, this phase does not produce a single deterministic trajectory. Instead, we leverage a \textit{conditional diffusion model} to model a multimodal joint \textit{trajectory distribution} over both the predicted future behaviors of heterogeneous entities and the task-directed trajectory plans of robots. During inference, the reverse diffusion process iteratively denoises a joint robot--entity trajectory sample, producing at each denoising step a clean estimate of the joint trajectories for subsequent safety-social guidance.

\noindent
$\bullet$ \textit{Phase 3: Safety-social guided reverse diffusion.}
At each reverse diffusion step, the clean joint trajectory estimate obtained in Phase 2 may violate safety, social, kinematic, or task-progress requirements. Phase 3 encodes these requirements as \textit{differentiable energies} and applies a bounded correction to the robot trajectory components while leaving the predicted entity trajectories unchanged. The corrected joint trajectory estimate is then re-noised to the next diffusion level and returned to Phase 2 for further denoising. This iterative denoising-guidance-re-noising cycle continues until the final diffusion step, yielding safety- and social-aware robot trajectory plans together with the corresponding entity trajectory predictions.

In the following, we define agents and trajectories (Secs. \ref{subsec:agent_modeling}--\ref{subsec:trajectory_modeling}), introduce perception neighborhoods and kinematic quantities (Sec.~\ref{Sec1-C}), and formulate the joint prediction--planning objective and safety-social constraints (Sec.~\ref{subsec:constraints}).

\vspace{-4.2mm}
\subsection{Agent Modeling in Heterogeneous Environments}
\label{subsec:agent_modeling}
\vspace{-.25mm}
We consider two key agent types: \textit{(i) environmental entities} such as heterogeneous PHAs, HNAAs, and SSAAs, which are dynamic and uncontrollable, gathered by the set $\mathcal{O}= \{o_1, \ldots, o_i,\ldots, o_{|\mathcal{O}|}\}$; and \textit{(ii) robots} that fall in the category of EAAs, collected by the set $\mathcal{R}=\{r_1, \ldots, r_j, \ldots, r_{|\mathcal{R}|}\}$. For each entity $o_i$ and robot $r_j$, we collect their semantic and geometric attributes in the following tuples:
\begin{equation}
\hspace{-2mm}\boldsymbol{a}_i^{(\mathsf{o})} = \left(\mathcal{C}_i^{(\mathsf{o,role})}, \mathcal{C}_i^{(\mathsf{o,geom})}\right),~\boldsymbol{a}_j^{(\mathsf{r})} = \left(\mathcal{C}_j^{(\mathsf{r,role})}, \mathcal{C}_j^{(\mathsf{r,geom})}\right),\hspace{-2mm}
\end{equation}
where $\mathcal{C}_i^{(\mathsf{o,role})} \in \mathcal{C}^{(\mathsf{role})}\setminus\{\mathsf{robot}\}   $ represents the semantic category (e.g., $\mathcal{C}^{(\mathsf{role})}=\{\mathsf{pedestrian, cyclist, vehicle,\ldots,robot}\}$), describing its behavioral patterns and social roles; and $\mathcal{C}_i^{(\mathsf{o,geom})},\mathcal{C}_j^{(\mathsf{r,geom})} \in \mathbb{R}^2$ represents the geometric properties (i.e., length and width) that define physical dimensions and collision boundaries. All robots share fixed geometric dimensions $\mathcal{C}_j^{(\mathsf{r,geom})}=\mathcal{C}^{(\mathsf{r,geom})}$,~$\forall j$, same role $\mathcal{C}_j^{(\mathsf{r,role})}=\mathcal{C}^{(\mathsf{r,role})} = \mathsf{robot}$,~$\forall j$, and kinematic constraints, while each robot $r_j$ is assigned a waypoint $\mathbf{g}_j=(x_j^{(\mathsf{g})},y_j^{(\mathsf{g})})$ as its navigation objective within the current prediction horizon $T^{(\mathsf{prd})}$.

% To standardize with the interaction representation, we assign all robots the same semantic category $\mathcal{C}^{(\mathsf{r,role})} = \mathsf{\{robot\}}$.

\vspace{-4.2mm}
\subsection{Trajectory Modeling}
\label{subsec:trajectory_modeling}
\vspace{-.25mm}
At each planning instant, SAGE processes a single observation window of length $T^{(\mathsf{obs})}$ and generates joint trajectories over a prediction horizon of length $T^{(\mathsf{prd})}$. Using a relative time index, $t=0$ denotes the current planning time, the observed time steps are $t \in \{-T^{(\mathsf{obs})} + 1, \ldots, 0\}$, and the future time steps are $t \in \{1, \ldots, T^{(\mathsf{prd})}\}$. The same procedure can be repeatedly invoked as new observations become available, while each planning instance is formulated independently.
For each entity $o_i\in \mathcal{O}$, we use the observed planar position $\mathbf{p}_i^{(\mathsf{o},t)}=(x_i^{(\mathsf{o},t)},y_i^{(\mathsf{o},t)})$ at time $t$ to construct the position history $\boldsymbol{s}_i^{(\mathsf{past})}=\{\mathbf{p}_i^{(\mathsf{o},t)}\mid t\in\{-T^{(\mathsf{obs})}+1,\ldots,0\}\}$. Velocity components can be obtained by finite differences when required. For each robot $r_j\in\mathcal{R}$,
 its planar position and its Cartesian state at time $t$ are $\mathbf{p}_j^{(\mathsf{r},t)}=(x_j^{(\mathsf{r},t)},y_j^{(\mathsf{r},t)})$ 
 and $\boldsymbol{s}_j^{(\mathsf{r},t)}=(x_j^{(\mathsf{r},t)},y_j^{(\mathsf{r},t)},v_j^{(\mathsf{rx},t)},v_j^{(\mathsf{ry},t)})$, respectively, while 
its current Cartesian state is given by $\boldsymbol{s}_j^{(\mathsf{r},0)}=(x_j^{(\mathsf{r},0)},y_j^{(\mathsf{r},0)},v_j^{(\mathsf{rx},0)},v_j^{(\mathsf{ry},0)})$, where $x_j^{(\mathsf{r},0)}$ and $y_j^{(\mathsf{r},0)}$ denote its current planar position coordinates, and $v_j^{(\mathsf{rx},0)}$ and $v_j^{(\mathsf{ry},0)}$ denote the velocity components along the $x$- and $y$-axes, estimated from the last two observed positions. Each robot $r_j$ aims to generate a trajectory from this current state toward its assigned waypoint $\mathbf{g}_j$ over $T^{(\mathsf{prd})}$, subject to safety and social-interaction requirements.

\vspace{-4.15mm}
\subsection{Perception Neighborhoods and Kinematic Quantities}\label{Sec1-C}
\vspace{-.25mm}

To facilitate constraint modeling, we next formalize the robot's local perception neighborhood (Sec.~\ref{SecIII-C1}) and relevant kinematic quantities (Sec.~\ref{SecIII-C2}). 

% Let $\mathbf{p}_j^{(\mathsf{r},t)}=(x_j^{(\mathsf{r},t)},y_j^{(\mathsf{r},t)})$ and $\mathbf{p}_i^{(\mathsf{o},t)}=(x_i^{(\mathsf{o},t)},y_i^{(\mathsf{o},t)})$ denote the planar positions of robot $r_j$ and neighbor $o_i$, respectively.

\vspace{-.1mm}
\subsubsection{Heterogeneity-aware neighborhood}\label{SecIII-C1}
In populated environments, exhaustively modeling all entities in $\mathcal{O}$ incurs substantial computational overhead and may introduce redundant or irrelevant interaction dependencies that degrade representation quality. Thus, for each robot $r_j \in \mathcal{R}$ at time $t$, we define its \textit{effective perception neighborhood} as the set of nearby entities within a pair-dependent interaction radius $R_{j,i}^{(\mathsf{val})}$ as follows:
\begin{equation}
\mathcal{N}_j^t =
\left\{
o_i
\;\middle|\;
o_i \in \mathcal{O},
\;
\left\|
\mathbf{p}_j^{(\mathsf{r},t)}
-
\mathbf{p}_i^{(\mathsf{o},t)}
\right\|_2
\le R_{j,i}^{(\mathsf{val})}
\right\},
\end{equation}
where $R_{j,i}^{(\mathsf{val})}$ is constrained by two factors: (i) the sensing range of robot $r_j$, which is bounded by its maximum observable range, and (ii) the category-dependent social influence range of neighbor $o_i$, which depends on its semantic role, motion state, and geometric properties. For instance, for pedestrians, we use Hall's proxemics theory~\cite{hall1966hidden} as a reference (e.g.,  the boundary between the social and public zones is approximately $3.6\,\mathrm{m}$). For other entity categories, such as cyclists or vehicles, the interaction radius can be adjusted according to their speed, size, and safety-relevant motion characteristics\footnote{Note that, $\mathcal{N}_j^t$ contains only heterogeneous non-robot neighbors observed by robot $r_j$ at time $t$, while nearby robots are later modeled separately through robot-to-robot (R2R) interactions.}.

\subsubsection{Kinematic quantities}\label{SecIII-C2}
To characterize the nonholonomic constraints of a robot, we define three derived quantities from its state vector $\boldsymbol{s}_j^{(\mathsf{r},t)}$ in the following. Let $\mathbf{v}_j^t=(v_j^{(\mathsf{rx},t)},v_j^{(\mathsf{ry},t)})$ denote the planar translational velocity of robot $r_j$ at time $t$.
First, the robot's \textit{heading angle} at time $t$ is given by:
\begin{equation}
\theta_j^t = \operatorname{atan2}\!\left(v_j^{(\mathsf{ry},t)}, v_j^{(\mathsf{rx},t)}+\varepsilon\right),
\end{equation}
where $0<\varepsilon\ll 1$ is a numerical stabilizer to avoid ambiguity in the case of $v_j^{(\mathsf{ry},t)}=v_j^{(\mathsf{rx},t)}=0$.
Subsequently, the \textit{lateral velocity} (i.e., the component of the current velocity perpendicular to the preceding heading) is given by:
\begin{equation}
v_j^{(\mathsf{lat},t)} =
-v_j^{(\mathsf{rx},t)} \sin(\theta_j^{t-1})
+ v_j^{(\mathsf{ry},t)} \cos(\theta_j^{t-1}).
\end{equation}
Finally, the \textit{angular velocity} is computed as follows:
\begin{equation}
\omega_j^t =
\begin{cases}
0, & t=1,\\
\dfrac{\operatorname{atan2}(\sin\Delta\theta_j^t,\cos\Delta\theta_j^t)}{\Delta t}, & t\ge 2,
\end{cases}
\end{equation}
where $\Delta\theta_j^t=\theta_j^t-\theta_j^{t-1}$,  and $\Delta t$ is the sampling interval (set to $0.4\,\mathrm{s}$ in our experiments). For $t=1$, $\theta_j^0$ estimated from the last two observations is used as the reference for $v_j^{(\mathsf{lat},1)}$. 

% For an ideal differential-drive robot, lateral motion should remain small and the translational and angular velocities should remain within their prescribed limits.

\vspace{-2mm}
\subsection{Optimization Objective and Safety-Social Constraints}
\label{subsec:constraints}
Next, we formulate the navigation problem of our interest with two coupled components. \textit{First}, we define a joint generative objective that maps current robot states, assigned waypoints, and historical observations of neighbor entities to a distribution over predicted entity trajectories and task-directed robot trajectory plans (Sec.~\ref{sebsubD1}). \textit{Second}, we impose task and safety-social feasibility constraints on samples from this distribution, including task progress, collision avoidance, kinematic feasibility, and social compliance (Sec.~\ref{sebsubD2}). 

\subsubsection{Joint generative objective}\label{sebsubD1}
Given the robot states at the current planning time (i.e., $\{\boldsymbol{s}_j^{(\mathsf{r},0)}\}_{r_j\in\mathcal{R}}$), their waypoints (i.e., $\{\mathbf{g}_j\}_{r_j\in\mathcal{R}}$), and the historical observations of the currently perceived heterogeneous entities (i.e., $\{\boldsymbol{s}_i^{(\mathsf{past})}\}_{o_i\in\mathcal{O}_{\mathcal{N}}^0}$), we define the \textit{scene-level conditioning input} as follows:
\begin{equation}
S^{(\mathsf{in})}
:=
\bigg\{
\bigcup_{r_j \in \mathcal{R}} \boldsymbol{s}_j^{(\mathsf{r},0)},
\;
\bigcup_{r_j \in \mathcal{R}} \mathbf{g}_j,
\;
\bigcup_{o_i \in \mathcal{O}_{\mathcal{N}}^0} \boldsymbol{s}_i^{(\mathsf{past})}
\bigg\}.
\end{equation}
% where  $\mathcal{O}_{\mathcal{N}}^0=\bigcup_{r_j\in\mathcal{R}}\mathcal{N}_j^0$ denotes the set of neighbor entities currently perceived by the robot team.
% The \textit{diffusion target} (i.e., the system output) is the joint future velocity sequence $S^{(\mathsf{out})}=\{S^{(\mathsf{r,out})},S^{(\mathsf{o,out})}\}$, where $S^{(\mathsf{r,out})}$ contains robot plan velocities and $S^{(\mathsf{o,out})}$ contains predicted entity velocities over $T^{(\mathsf{prd})}$. Positions used for navigation and energy evaluation are recovered by $\mathbf{p}_a^t=\mathbf{p}_a^0+\Delta t\sum_{\tau=1}^{t}\mathbf{v}_a^\tau$ for each robot or entity $a$.
% Note that trajectory prediction and planning are fundamentally coupled. Without anticipating neighboring agents, navigation reduces to open-loop planning in a static environment and cannot account for the reciprocal, evolving dynamics of multi-agent interactions. We therefore formulate them as a unified problem and use a \textit{conditional generative model} to parameterize the \textit{joint trajectory distribution} $p_{\theta}(S^{(\mathsf{out})}\mid S^{(\mathsf{in})})$, where $\theta$ denotes the learnable parameters of the scene encoder and the diffusion denoising network. The model parameters $\theta$ are learned by minimizing the diffusion reconstruction objective $\mathcal{L}_{\mathsf{diff}}$ (later detailed in Sec.~\ref{subsec:phase2}).
where $\mathcal{O}_{\mathcal{N}}^0=\bigcup_{r_j\in\mathcal{R}}\mathcal{N}_j^0$ denotes the set of neighbor entities currently perceived by the robot team. Given $S^{(\mathsf{in})}$, our objective is to jointly generate (i) the future trajectories of the perceived heterogeneous entities and (ii) task-directed trajectories for the robots. Accordingly, we define the system output as the joint future velocity set
$
S^{(\mathsf{out})}
=
\left\{
S^{(\mathsf{r,out})},
S^{(\mathsf{o,out})}
\right\},
$
where $S^{(\mathsf{r,out})}$ contains the planned robot velocities and $S^{(\mathsf{o,out})}$ contains the predicted entity velocities over the prediction horizon $T^{(\mathsf{prd})}$. 
Since $S^{(\mathsf{out})}$ consists of velocity sequences, the future robot and entity positions can be obtained by integrating their velocities as
$
\mathbf{p}_j^{(\mathsf{r},t)}
{=}
\mathbf{p}_j^{(\mathsf{r},0)}
+
\Delta t\sum_{\tau=1}^{t}\mathbf{v}_j^{\tau},~~
\mathbf{p}_i^{(\mathsf{o},t)}
{=}
\mathbf{p}_i^{(\mathsf{o},0)}
+
\Delta t\sum_{\tau=1}^{t}\mathbf{v}_i^{\tau}.
$

% The corresponding positions are recovered as
% $\mathbf{p}_a^t=\mathbf{p}_a^0+\Delta t\sum_{\tau=1}^{t}\mathbf{v}_a^\tau$
% for each robot or entity $a$.

Since entity trajectory prediction and robot trajectory planning are inherently coupled, we model their joint output through the conditional distribution
$
p_{\theta}\!\left(
S^{(\mathsf{out})}
\mid
S^{(\mathsf{in})}
\right),
$
where $\theta$ denotes learnable parameters. This distribution captures the multimodal future behaviors of heterogeneous entities together with the  task-directed robot trajectory plans. In Sec.~\ref{subsec:phase2}, we realize this conditional distribution using a diffusion-based generative model, whose parameters (i.e., $\theta$) are learned by minimizing the diffusion reconstruction objective $\mathcal{L}_{\mathsf{diff}}$.

\subsubsection{Safety-social constrained trajectory planning}\label{sebsubD2}
Although samples from the learned conditional joint trajectory distribution $p_{\theta}(S^{(\mathsf{out})}\mid S^{(\mathsf{in})})$ are statistically plausible, they may fail to make task progress or violate safety, kinematic, or social requirements, especially in dense/long-tail interaction scenarios. Hence, we seek robot trajectories that are both likely under this distribution and feasible with respect to task, physical, and social constraints, leading to our problem formulation $\bm{\mathcal{P}}$:
\begin{equation} \label{eq:MainPro}
\begin{aligned}
(\bm{\mathcal{P}}): \quad
&
S^{(\mathsf{out})}
\sim
p_{\theta}
\left(
\cdot
\mid
S^{(\mathsf{in})}
\right)
\quad
\text{\textbf{s.t.}}
\quad
\mathrm{(C1)}\text{--}\mathrm{(C5)}.
\end{aligned}
\end{equation}
Specifically, constraints (C1)-(C5) are detailed below.

\noindent
\textbullet~\textit{(C1) Robot-to-neighbor collision avoidance.}
Each robot should maintain its pairwise safety margin from neighbors:
\begin{equation}
\begin{alignedat}{2}
\left\|
\mathbf{p}_j^{(\mathsf{r},t)}
-
\mathbf{p}_i^{(\mathsf{o},t)}
\right\|_2
&\ge m_{j,i}, \\
\forall t \in \{1,\ldots,T^{(\mathsf{prd})}\},\ 
&\forall r_j \in \mathcal{R},\ 
\forall o_i \in \mathcal{N}_j^t,
\end{alignedat}
\end{equation}
where $m_{j,i}$  denotes the safety margin between robot $r_j$ and entity neighbor $o_i$.

\noindent
\textbullet~\textit{(C2) Robot-to-robot (R2R) collision avoidance.}
Any pair of robots should preserve their pairwise safety margin:
\begin{equation}
\begin{alignedat}{2}
\left\|
\mathbf{p}_j^{(\mathsf{r},t)}
-
\mathbf{p}_{j'}^{(\mathsf{r},t)}
\right\|_2
&\ge m_{j,j'}, \\
\forall t \in \{1, \ldots, T^{(\mathsf{prd})}\},\ 
&\forall r_j,r_{j'} \in \mathcal{R},\ 
j' \neq j,
\end{alignedat}
\end{equation}
where $m_{j,j'}$  is the safety margin between robots $r_j$ and $r_{j'}$.

% Here, $d^{(\mathsf{safe})}$ is the base clearance threshold. When geometry-aware margins are enabled, $\mathcal C_a^{(\mathsf{geom})}$ denotes the length--width tuple of agent $a$, and $\frac12\|\mathcal C_a^{(\mathsf{geom})}\|_2$ is the radius of its circumscribed circle; thus, for any interacting pair $a,b$, $m_{a,b}=d^{(\mathsf{safe})}+\frac12\|\mathcal C_a^{(\mathsf{geom})}\|_2+\frac12\|\mathcal C_b^{(\mathsf{geom})}\|_2$. Otherwise, agents are treated as points for margin calculation and $m_{a,b}=d^{(\mathsf{safe})}$.

In the above constraints (C1) and (C2), the safety margins $m_{j,i}$ and $m_{j,j'}$ account for both a prescribed minimum clearance $d^{(\mathsf{safe})}$ and the physical dimensions of the interacting agents. 
Specifically, approximating each robot and entity by its circumscribed circle, the safety margins are given by
$
m_{j,i}
=
d^{(\mathsf{safe})}
+\frac{1}{2}\|\mathcal{C}_j^{(\mathsf{r,geom})}\|_2
+\frac{1}{2}\|\mathcal{C}_i^{(\mathsf{o,geom})}\|_2,
$
and
$
m_{j,j'}
=
d^{(\mathsf{safe})}
+\frac{1}{2}\|\mathcal{C}_j^{(\mathsf{r,geom})}\|_2
+\frac{1}{2}\|\mathcal{C}_{j'}^{(\mathsf{r,geom})}\|_2
$.

\noindent
\textbullet~\textit{(C3) Kinematic feasibility.}
The generated robot trajectories should satisfy the prescribed kinematic limits: 
\begin{equation}
\begin{gathered}
\|\mathbf{v}_j^t\|_2\le v_{\max},\quad
|v_j^{(\mathsf{lat},t)}|\le v_{\max}^{\mathsf{lat}},\quad
|\omega_j^t|\le \omega_{\max},\\[-1mm]
\forall t \in \{1, \ldots, T^{(\mathsf{prd})}\},\ \forall r_j \in \mathcal{R}.
\end{gathered}
\end{equation}

\noindent
\textbullet~\textit{(C4) Social compliance.}
Beyond physical collision avoidance, robot trajectories should remain outside the category-conditioned social compliance region induced by neighbors in the robot's forward half-plane:
\begin{equation}\label{eq:outside-SC}
\begin{aligned}
\left(
\frac{d_{i,j}^{(\mathsf{lon},t)}}
{\sigma^{\mathsf{lon}}\!\left(\mathcal{C}_i^{(\mathsf{o,role})}\right)}
\right)^2
+
\left(
\frac{d_{i,j}^{(\mathsf{lat},t)}}
{\sigma^{\mathsf{lat}}\!\left(\mathcal{C}_i^{(\mathsf{o,role})}\right)}
\right)^2
\ge 1,
\\
\forall t \in \{1, \ldots, T^{(\mathsf{prd})}\},\ \forall r_j \in \mathcal{R},\ \forall o_i \in \mathcal{N}_j^t, d_{i,j}^{(\mathsf{lon},t)}\geq 0.
\end{aligned}
\end{equation}
% In~\eqref{eq:outside-SC}, $d_{i,j}^{(\mathsf{lon},t)}$ and $d_{i,j}^{(\mathsf{lat},t)}$ can be obtained by projecting the relative position from robot $r_j$ to neighbor $o_i$ into the robot-centric frame:
In~\eqref{eq:outside-SC}, $d_{i,j}^{(\mathsf{lon},t)}$ and $d_{i,j}^{(\mathsf{lat},t)}$ denote the longitudinal and lateral components, respectively, of the position of neighbor $o_i$ with respect to robot $r_j$, expressed in the robot-centric coordinate:
\begin{equation}
\begin{bmatrix}
d_{i,j}^{(\mathsf{lon},t)} \\
d_{i,j}^{(\mathsf{lat},t)}
\end{bmatrix}
=
\begin{bmatrix}
\cos(\theta_j^t) & \sin(\theta_j^t) \\
-\sin(\theta_j^t) & \cos(\theta_j^t)
\end{bmatrix}
\begin{bmatrix}
x_i^{(\mathsf{o},t)} - x_j^{(\mathsf{r},t)} \\
y_i^{(\mathsf{o},t)} - y_j^{(\mathsf{r},t)}
\end{bmatrix}.
\end{equation}
% Also, the functions $\sigma^{\mathsf{lon}}(\cdot)$ and $\sigma^{\mathsf{lat}}(\cdot)$ define the role-conditioned semi-axes of the elliptical comfort region.
Also, $\sigma^{\mathsf{lon}}(\cdot)$ and $\sigma^{\mathsf{lat}}(\cdot)$ define the longitudinal and lateral extents of the elliptical comfort region, respectively, based on the neighbor's semantic role (e.g., $\mathcal{C}_i^{(\mathsf{o,role})}\in \mathsf{\{pedestrian, cyclist, vehicle\}}$).
In words, for each forward neighbor (imposed by $d_{i,j}^{(\mathsf{lon},t)}\geq 0$),~\eqref{eq:outside-SC} requires the robot to remain on or outside this ellipse. 
The right-hand side of~\eqref{eq:outside-SC} is set to one because the longitudinal and lateral separations are normalized by their corresponding role-dependent comfort-region extents. Thus, values below one indicate intrusion into the comfort region, whereas values greater than or equal to one satisfy the social-compliance constraint. These role-dependent margins provide an interpretable social prior, with larger margins for pedestrians reflecting interpersonal comfort and smaller margins for vehicles  capturing safety considerations.

% The right-hand side of~\eqref{eq:outside-SC} is set to one because the separations are normalized by these semi-axes: values below one indicate intrusion, whereas values at or above one satisfy the constraint; any other positive threshold would merely rescale the margins. Shared category-level margins provide an interpretable social prior, with larger pedestrian margins reflecting interpersonal comfort and tighter vehicle margins primarily reflecting safety.

\noindent
\textbullet~\textit{(C5) Task progress.}
The generated robot trajectory should reach the assigned waypoint $\mathbf{g}_j$ or terminate sufficiently close to it within the planning horizon, which results in 
\begin{equation}\label{eq:endSeg}
\begin{aligned}
\left\|
\mathbf{p}_j^{(\mathsf{r},T^{(\mathsf{prd})})}
-
\mathbf{g}_j
\right\|_2
\le d^{(\mathsf{goal})},
\quad
\forall r_j \in \mathcal{R},
\end{aligned}
\end{equation}
where $d^{(\mathsf{goal})}$ is the tolerance of vicinity to the waypoint.

The formulation $\bm{\mathcal{P}}$ in~\eqref{eq:MainPro} specifies both the generative target and the task-feasibility requirements of socially-aware robot navigation. Nevertheless, directly solving $\bm{\mathcal{P}}$ is fundamentally challenging due to the intricate coupling among multimodal entity trajectory prediction and robot trajectory planning, heterogeneous interaction patterns of agents, collision avoidance, kinematic feasibility, and socially compliant behavior modeling. These are inherently interdependent and often exhibit competing constraints across spatial, temporal, and social dimensions. To address this complexity, we next develop SAGE.

\vspace{-2mm}
\section{SAGE: A Socially-Aware Generative Engine}
% Building on Sec.~\ref{subsec:constraints}, this section presents how SAGE parameterizes and subsequently samples from the task-aware generative prior $p_\theta(S^{(\mathsf{out})}\mid S^{(\mathsf{in})})$. During training, SAGE learns this prior to capture the multimodal joint future distribution of controllable robots and heterogeneous entities within their perception neighborhoods. During inference, samples from this prior are further refined by safety-social energy guidance to promote task progress, collision avoidance, kinematic feasibility, and social compliance while remaining likely under the learned distribution.
% Specifically, SAGE consists of three synergistic phases. \textit{First}, an HGT encodes asymmetric robot-entity interactions into a contextual representation $C^{(\mathsf{ctx})}$ that captures social semantics and interaction hierarchies (\textit{Phase 1}). \textit{Second}, conditioned on $C^{(\mathsf{ctx})}$, a diffusion model jointly generates robot and entity trajectories, enabling coherent reasoning over coupled multi-agent dynamics (\textit{Phase 2}). \textit{Third}, differentiable safety-social energy guidance is integrated into the denoising process to reduce physical-safety and social-compliance energy penalties (\textit{Phase 3}). The overall  procedure of SAGE is summarized in Algorithm~\ref{alg:sage}.

Building on Sec.~\ref{subsec:constraints}, this section presents how SAGE parameterizes and subsequently samples from the task-aware generative prior $p_\theta(S^{(\mathsf{out})}\mid S^{(\mathsf{in})})$. During training, SAGE learns this prior to capture the multimodal joint future distribution of controllable robots and heterogeneous entities within their perception neighborhoods. During inference, SAGE samples from this prior through an iterative reverse diffusion process, during which safety-social energy guidance is applied to promote task progress, collision avoidance, kinematic feasibility, and social compliance while remaining likely under the learned distribution. 
Specifically, SAGE consists of three synergistic phases. \textit{First}, an HGT encodes asymmetric robot-entity interactions into a contextual representation $C^{(\mathsf{ctx})}$ that captures social semantics and interaction patterns (\textit{Phase 1}). \textit{Second}, conditioned on $C^{(\mathsf{ctx})}$, a diffusion model iteratively denoises a joint robot--entity trajectory sample, enabling coherent reasoning over coupled multi-agent dynamics (\textit{Phase 2}). \textit{Third}, at each reverse diffusion step, differentiable safety-social energy guidance corrects the denoised robot trajectories before proceeding to the next diffusion step, reducing physical-safety and social-compliance energy penalties (\textit{Phase 3}). The overall procedure of SAGE is summarized in Alg.~\ref{alg:sage} and its three phases are elaborated in the following.

\vspace{-2mm}
\subsection{Phase 1: Scene Encoding and Interaction Modeling}
\label{subsec:phase1}

% This phase encodes the heterogeneous scene described in Sec.~\ref{subsec:agent_modeling} as a context-token sequence $C^{(\mathsf{ctx})}$, providing the diffusion model with structured awareness of entity identities, categories, and social interaction patterns. To this end, as explained in the following, the scene is formulated as a directed heterogeneous graph, in which information passing is conducted via a tailored HGT to yield socially enriched node embeddings. Robot-centered interaction relationships are then captured as a directed heterogeneous graph representation:
% \begin{equation}
% \label{eq:heterogeneous_graph}
% \mathcal{G} = (\mathcal{V}, \mathcal{E}),
% \end{equation}
% where $\mathcal{V}=\mathcal{R}\cup\mathcal{O}_{\mathcal{N}}^0$ uses the team-level perceived entity set defined in Sec.~\ref{subsec:constraints}; under full-observability simulation, all active entities are treated as perceived. The encoder connects every robot--entity pair in this set through directed robot-to-neighbor (R2N) and neighbor-to-robot (N2R) edges, and all ordered robot pairs, including robot self-attention, through robot-to-robot (R2R) edges. We next construct the context-token sequence in three steps.

This phase transforms/encodes the heterogeneous scene described in Sec.~\ref{subsec:agent_modeling} into a \textit{context-token sequence} $C^{(\mathsf{ctx})}$ that conditions the diffusion model on agent-specific attributes and interaction patterns. 
To capture the role-dependent and asymmetric nature of agent interactions, we represent the observed scene as a \textit{directed heterogeneous graph}:
\begin{equation}
\label{eq:heterogeneous_graph}
\mathcal{G} = (\mathcal{V}, \mathcal{E}),
\end{equation}
where $\mathcal{V}=\mathcal{R}\cup\mathcal{O}_{\mathcal{N}}^0$ comprises the robots and currently perceived entities defined in Sec.~\ref{subsec:constraints}.  The edge set $\mathcal{E}$ captures directed robot-to-neighbor (R2N), neighbor-to-robot (N2R), and robot-to-robot (R2R) interactions, with R2R edges defined over all ordered robot pairs, including self-loops; the direction of each edge allows opposite interaction directions (e.g., R2N vs. N2R) to be modeled explicitly.
We next construct the context-token sequence $C^{(\mathsf{ctx})}$ in three steps, where a tailored HGT performs relation-specific information exchange over $\mathcal{G}$ to produce interaction-aware node embeddings.

\begin{algorithm}[t]
\caption{Workflow of SAGE}
\label{alg:sage}
{\footnotesize
\KwIn{Scene-level conditioning input $S^{(\mathsf{in})}$, robot and entity attributes, diffusion steps $K$, sampling stride $s$}
\KwOut{Robot trajectory plans $S^{(\mathsf{r,out})}$ and entity trajectory predictions $S^{(\mathsf{o,out})}$}
Construct the heterogeneous graph $\mathcal{G}$ with R2N, N2R, and R2R edges following~\eqref{eq:heterogeneous_graph}\;
Encode node features and obtain context tokens $C^{(\mathsf{ctx})}$ following~\eqref{eq:entity_node_embedding}--\eqref{eq:scene_context}\;
Initialize $S_K^{(\mathsf{out})}\sim\mathcal{N}(0,I)$ as the large-$K$ limit of~\eqref{eq:forward_closed_form}\;
\For{$k=K,K-s,\ldots$ while $k>0$}{
Predict the noise $\epsilon_\theta(S_k^{(\mathsf{out})},\beta_k,C^{(\mathsf{ctx})})$ using the denoiser trained via~\eqref{eq:diffusion_loss}\;
Estimate the clean velocity sequence $\hat{S}_k^{(\mathsf{out})}$ following~\eqref{eq:reverse_diffusion}\;
Compute the four energy terms following~\eqref{eq:total_energy}--\eqref{eq:goal_energy}\;
Apply the bounded robot-only correction following~\eqref{eq:guided_correction}\;
Set $k'=\max(k-s,0)$ and sample $S_{k'}^{(\mathsf{out})}$ following~\eqref{eq:guided_correction}\;
}
\Return{$S^{(\mathsf{r,out})}$ as robot trajectory plans and $S^{(\mathsf{o,out})}$ as entity trajectory predictions}
}
\end{algorithm}

\noindent
\textit{$\bullet$ Step 1: Node feature embedding.}
Before performing information exchange over $\mathcal{G}$, we construct a \textit{layer-0 embedding} for each entity and robot node. Since entity motion is inferred from historical observations, whereas robot planning depends on the current state and assigned waypoint, we use a long short-term memory (LSTM) network to encode entity trajectories and a multilayer perceptron (MLP) to encode robot states. 
These motion representations are then combined with semantic role, control status, geometry, and waypoint information to obtain node embeddings in a common latent space. Specifically, we concatenate ($\oplus$) these features and pass them through a fusion projection $\Phi_{\mathsf{fuse}}$, which consists of layer normalization followed by an MLP. Accordingly, for each entity $o_i\in\mathcal{O}_{\mathcal{N}}^0$, the layer-0 embedding is given by:
\begin{equation}
\label{eq:entity_node_embedding}
\begin{aligned}
\mathbf{h}_i^{(\mathsf{o},0)}=\Phi_{\mathsf{fuse}}\Big(
\mathrm{LSTM}(\boldsymbol{s}_i^{(\mathsf{past})})
\oplus \mathrm{Emb}_{\mathsf{role}}(\mathcal{C}_i^{(\mathsf{o,role})})
\\
\oplus \mathrm{Emb}_{\mathsf{ctrl}}(0)
\oplus \mathrm{MLP}_{\mathsf{geom}}(\mathcal{C}_i^{(\mathsf{o,geom})})
\oplus \mathrm{MLP}_{\mathsf{goal}}(\mathbf{0})\Big),
\end{aligned}
\end{equation}
where the LSTM encodes the observed position history $\boldsymbol{s}_i^{(\mathsf{past})}$ into its final hidden state, while $\mathrm{Emb}_{\mathsf{role}}(\cdot)$ denotes a learnable embedding that maps the entity's semantic role (e.g., pedestrian, cyclist, or vehicle) to a latent representation, and $\mathrm{MLP}_{\mathsf{geom}}(\cdot)$ encodes its physical dimensions. Moreover, $\mathrm{Emb}_{\mathsf{ctrl}}(0)$ identifies the entity as uncontrollable, while the zero waypoint input (i.e., $\mathrm{MLP}_{\mathsf{goal}}(\mathbf{0})$) indicates that no task objective is assigned to entity nodes.
Similarly, for each robot $r_j\in\mathcal{R}$, the layer-0 embedding is given by:
\begin{equation}\label{eq:robot_node_embedding}
\hspace{-3mm}
\hspace{-3mm}
\resizebox{0.455\textwidth}{!}{$
\begin{aligned}
&\mathbf{h}_j^{(\mathsf{r},0)}{=}\Phi_{\mathsf{fuse}}\Big(
\mathrm{MLP}_{\mathsf{state}}(\boldsymbol{s}_j^{(\mathsf{r},0)})
\oplus \mathrm{Emb}_{\mathsf{ctrl}}(1)\oplus \mathrm{Emb}_{\mathsf{role}}(\mathcal{C}^{(\mathsf{r,role})})\\[-1mm]
&
\oplus \mathrm{MLP}_{\mathsf{geom}}(\mathcal{C}^{(\mathsf{r,geom})})\oplus \mathrm{MLP}_{\mathsf{goal}}\big(
\mathbf{g}_j-\mathbf{p}_j^{(\mathsf{r},0)}
\big)\Big),
\end{aligned}
$}
\hspace{-6.5mm}
\end{equation}
where $\mathrm{MLP}_{\mathsf{state}}(\cdot)$ encodes the robot's current position and velocity, $\mathrm{Emb}_{\mathsf{ctrl}}(1)$ identifies the robot as controllable, and $\mathrm{MLP}_{\mathsf{goal}}(\cdot)$ encodes the displacement from its current position to the assigned waypoint. In~\eqref{eq:robot_node_embedding}, the role and geometry features are encoded analogously to those of the entity nodes.

The resulting layer-0 embeddings provide a common representation of motion, semantic role, control status, physical geometry, and task information, and are subsequently processed by the relation-specific HGT in Step~2. All components of the scene encoder, including the LSTM, embeddings, feature MLPs, fusion projection, and HGT, are jointly trained with the diffusion denoiser through $\mathcal{L}_{\mathsf{diff}}$ (later detailed in Sec.~\ref{subsec:phase2}).

\noindent
\textit{$\bullet$ Step 2: Meta-relation-based attention mechanism.}
For notational convenience, let $u_a,u_b\in\mathcal{V}$ denote arbitrary nodes in $\mathcal{G}$ defined in \eqref{eq:heterogeneous_graph}, where each node represents either a robot in $\mathcal{R}$ or a perceived entity in $\mathcal{O}_{\mathcal{N}}^0$. We use $\mathbf{h}_a^{(0)}$ to denote the layer-0 embedding of node $u_a$, obtained from Step~1. Specifically,
\begin{equation}
\mathbf{h}_a^{(0)}
=
\begin{cases}
\mathbf{h}_i^{(\mathsf{o},0)}, & \text{if}~u_a=o_i\in\mathcal{O}_{\mathcal{N}}^0,\\
\mathbf{h}_j^{(\mathsf{r},0)}, & \text{if}~u_a=r_j\in\mathcal{R}.
\end{cases}
\end{equation}
These layer-0 embeddings serve as the initial node representations processed by the HGT. As discussed in the following, the subsequent HGT layers then produce updated representations $\mathbf{h}_a^{(l)}$, $l=1,\ldots,L$, by exchanging information over the directed edges of $\mathcal{G}$.
To this end, to capture asymmetric, role-dependent interactions among robots and heterogeneous entities, we assign a distinct relation type to each directed edge in $\mathcal{G}$ based on the semantic roles of its source and target nodes.\footnote{A Graph Attention Network (GAT)~\cite{velickovic2018graph} typically applies shared attention parameters across edges and therefore does not explicitly distinguish heterogeneous, direction-dependent interaction types. In contrast, our setting requires different parameterizations for ordered role pairs (e.g., robot-to-pedestrian versus pedestrian-to-robot), motivating the relation-specific HGT formulation.}
 Specifically, for a directed edge $u_b\rightarrow u_a$, with $u_b$ and $u_a$ denoting the source and target nodes, respectively, we define the relation ID as:
\begin{equation}
\psi(u_b\!\rightarrow u_a)
=\iota(u_b)N_{\mathsf{role}}+\iota(u_a),
\end{equation}
where $\iota(u)\in\{0,\ldots,N_{\mathsf{role}}-1\}$ denotes the semantic-role ID of node $u$, and $N_{\mathsf{role}}$ is the total number of semantic roles. Since the source and target roles are ordered, reversing an edge generally results in a different relation ID. This allows SAGE to distinguish asymmetric interactions, such as the influence of a pedestrian on a robot from that of a robot on a pedestrian.

Based on these relation types, the HGT iteratively updates each node by attending to and aggregating information from its incoming neighbors. Specifically, for a target node $u_a$, let
\begin{equation}
\mathcal{M}_a=\{u_b\in\mathcal{V}\mid(u_b\rightarrow u_a)\in\mathcal{E}\}
\end{equation}
denote its set of incoming source nodes. Each incoming edge $u_b\rightarrow u_a$ is associated with a relation ID
$\psi(a,b):=\psi(u_b\rightarrow u_a)$.
To capture different aspects of an interaction, the HGT employs \textit{multi-head attention} with $H$ parallel attention heads, indexed by $h\in\{1,\ldots,H\}$. Each attention head independently evaluates the relevance of incoming source nodes and extracts the information they contribute to the target node. To further allow various interaction types to be processed differently, each relation type $\psi(a,b)$ and attention head $h$ are associated with learnable/trainable \textit{query}, \textit{key}, and \textit{value} projection matrices
$(W_{\psi(a,b),h}^{\mathsf{Q}},W_{\psi(a,b),h}^{\mathsf{K}},W_{\psi(a,b),h}^{\mathsf{V}})$,
along with an \textit{attention-logit bias} $b_{\psi(a,b),h}$. The query and key projections transform the target and source node representations into a common space in which their interaction relevance is measured, while the value projection transforms the source-node representation into the information to be passed to the target node. Thus, within each attention head, the relation ID $\psi(a,b)$ selects the set of learnable parameters used to determine both \textit{how relevant} source node $u_b$ is to target node $u_a$ and \textit{what information} $u_b$ contributes to $u_a$.
Subsequently, at HGT layer $l\in\{1,\ldots,L\}$, the attention score between source node $u_b$ and target node $u_a$, and its normalized attention weight, are given by:
\begin{equation}
\hspace{-3mm}
\resizebox{0.44\textwidth}{!}{$
\begin{aligned}
e_{a,b,h}^{(l)} &=
\frac{
(\mathbf{h}_a^{(l-1)}W_{\psi(a,b),h}^{\mathsf{Q}})
(\mathbf{h}_b^{(l-1)}W_{\psi(a,b),h}^{\mathsf{K}})^\top
}{\sqrt{d_h}}
+b_{\psi(a,b),h}~\hspace{-.8mm},\\
\alpha_{a,b,h}^{(l)} &=
\operatorname{Softmax}_{u_b\in\mathcal{M}_a}
\big(e_{a,b,h}^{(l)}\big),
\end{aligned}
$}
\hspace{-3mm}
\end{equation}
where $d_h$ denotes the per-head projection dimension. In essence, the attention weight $\alpha_{a,b,h}^{(l)}$ determines the relative importance of source node $u_b$ to target node $u_a$ at layer $l$. Using these attention weights, each attention head aggregates the value-projected representations of all incoming source nodes. We refer to this weighted aggregate as the \textit{head-wise message} $\mathbf{m}_{a,h}^{(l)}$ received by target node $u_a$, which is given by:
\begin{equation}
\mathbf{m}_{a,h}^{(l)}
=
\sum_{u_b\in\mathcal{M}_a}
\alpha_{a,b,h}^{(l)}
\mathbf{h}_b^{(l-1)}
W_{\psi(a,b),h}^{\mathsf{V}}.
\end{equation}
The $H$ head-wise messages are then concatenated and modulated by a \textit{relation-dependent gate} before being used to update the representation of target node $u_a$ as:
\begin{equation}
\hspace{-2mm}\begin{aligned}
\mathbf{h}_a^{(l)}
=\rho_{\mathsf{HGT}}\!\Bigl(
\mathbf{h}_a^{(l-1)},
W^{\mathsf O}\!\left[
\boldsymbol{\eta}_a^{(l)}\odot
\operatorname{Concat}_{h=1}^{H}
\left(\mathbf{m}_{a,h}^{(l)}\right)
\right]\Bigr),
\end{aligned}
\hspace{-2mm}
\end{equation}
where $\boldsymbol{\eta}_a^{(l)}
=\operatorname{sigmoid}\!\Big(
\frac{1}{\max(1,|\Psi_a|)}
\sum_{\psi\in\Psi_a}\boldsymbol{\gamma}_{\psi}
\Big)$
is the relation-dependent gate that modulates the aggregated messages, with
$\Psi_a=\{\psi(u_b\rightarrow u_a)\mid u_b\in\mathcal{M}_a\}$ denoting the set of relation types associated with the incoming edges of $u_a$ and $\boldsymbol{\gamma}_{\psi}$ denoting the learnable gate associated with relation $\psi$; $\operatorname{Concat}_{h=1}^{H}(\cdot)$ concatenates the $H$ head-wise messages; $\odot$ denotes element-wise multiplication; $W^{\mathsf O}$ is a learnable output projection that maps the concatenated multi-head representation back to the node-embedding space; and $\rho_{\mathsf{HGT}}$ applies the residual connection, layer normalization, and GELU-based feed-forward update. Repeating this process over $L$ HGT layers enables each node to progressively incorporate role-dependent information from its multi-hop interaction neighborhood. The final representation $\mathbf{h}_a^{(L)}$ therefore constitutes the learned interaction-aware embedding of node $u_a$ produced by the HGT.

\noindent
\textit{$\bullet$ Step 3: Scene context construction.}
After $L$ HGT layers, the final interaction-aware node representations $\mathbf{h}_a^{(L)}$ are retained as individual node tokens. To additionally capture global scene-level information, we aggregate the final representations of all robot and entity nodes through mean pooling (i.e., element-wise averaging) to obtain a graph-level token $\mathbf{c}^{(\mathsf{G})}
=
\operatorname{Norm}\!\left(
\Phi_{\mathsf{ctx}}\!\left(
\frac{1}{|\mathcal{V}|}
\sum_{a=1}^{|\mathcal{V}|}
\mathbf{h}_a^{(L)}
\right)
\right)$, where $\Phi_{\mathsf{ctx}}$ is a learnable projection that maps the pooled graph representation into the node-token dimension and $\operatorname{Norm}(\cdot)$ denotes $\ell_2$ normalization. 
Finally, the context-token sequence is constructed as follows:
\begin{equation}
\label{eq:scene_context}
C^{(\mathsf{ctx})}
=
\left[
\mathbf{h}_1^{(L)},\ldots,
\mathbf{h}_{|\mathcal{V}|}^{(L)},
\mathbf{c}^{(\mathsf{G})}
\right].
\end{equation}
  Consequently, $C^{(\mathsf{ctx})}$ provides the diffusion model in Phase~2 with both agent-specific interaction representations and a global representation of the overall scene.

% After $L$ HGT layers, we retain the node tokens and append a normalized mean-pooled graph token to construct the context-token sequence as:
% \begin{equation}
% \label{eq:scene_context}
% \begin{aligned}
% C^{(\mathsf{ctx})}
% &=\left[\mathbf h_{u_1}^{(L)},\ldots,\mathbf h_{u_{|\mathcal V|}}^{(L)},\mathbf c^{(\mathsf G)}\right].
% \end{aligned}
% \end{equation}
% where $\mathbf{c}^{(\mathsf G)}
% =\operatorname{Norm}\!\left(\Phi_{\mathsf{ctx}}\!\left(
% \frac{1}{|\mathcal V|}\sum_{u\in\mathcal V}\mathbf h_u^{(L)}\right)\right)$
% is XXXXXX
% and  $\Phi_{\mathsf{ctx}}$ projects the pooled feature into the node-token dimension and $\operatorname{Norm}$ denotes $\ell_2$ normalization. Thus, $C^{(\mathsf{ctx})}$ provides Phase~2 with agent-specific interaction tokens and a global scene summary. HGT supplies an implicit interaction prior for joint generation, whereas Phase~3 provides explicit constraint-aware correction.

\vspace{-3mm}
\subsection{Phase 2: Conditional Diffusion Trajectory Generation}
\label{subsec:phase2}

Conditioned on the context tokens $C^{(\mathsf{ctx})}$ obtained in Phase~1, Phase~2 employs a denoising diffusion probabilistic model (DDPM)~\cite{ho2020denoising} to model the conditional joint trajectory distribution
$p_\theta\!\left(S^{(\mathsf{out})}\mid C^{(\mathsf{ctx})}\right)$.
By sampling from this distribution, SAGE jointly generates robot plan velocities and entity velocity predictions, thereby capturing their coupled and multimodal future dynamics. We next describe the forward diffusion process, context-conditioned denoiser and training objective, and reverse sampling process underlying our DDPM.

\noindent
$\bullet$ \textit{Forward diffusion process.}
During training, each data sample consists of a scene-level conditioning input $S^{(\mathsf{in})}$ (defined in Sec.~\ref{sebsubD1}) and its corresponding ground-truth joint future velocity sequence $S^{(\mathsf{out})}$. Phase~1 maps $S^{(\mathsf{in})}$ to the context tokens $C^{(\mathsf{ctx})}$, whereas the forward diffusion process progressively adds Gaussian noise to  $S^{(\mathsf{out})}$ over $K$ diffusion steps. Specifically, at diffusion step $k\in\{1,\ldots,K\}$, the transition from $S_{k-1}^{(\mathsf{out})}$ to $S_k^{(\mathsf{out})}$ is defined as follows:
\begin{equation}
\label{eq:forward_diffusion}
q\!\left(S_k^{(\mathsf{out})}\mid S_{k-1}^{(\mathsf{out})}\right)
=
\mathcal{N}\!\left(
\sqrt{\alpha_k}S_{k-1}^{(\mathsf{out})},
\beta_k I
\right),
\end{equation}
where $S_0^{(\mathsf{out})}=S^{(\mathsf{out})}$ is the clean ground-truth joint future velocity sequence, $\beta_k$ is the prescribed noise variance at diffusion step $k$, and $\alpha_k=1-\beta_k$. Thus, at each forward diffusion step, $S_k^{(\mathsf{out})}$ is sampled from a Gaussian distribution with mean $\sqrt{\alpha_k}S_{k-1}^{(\mathsf{out})}$ and covariance $\beta_k I$, thereby progressively perturbing the joint future velocity sequence of robots and entities with Gaussian noise.
Rather than sequentially applying all preceding diffusion steps, a noisy sequence at any diffusion step $k$ can be sampled directly from the clean sequence as:
\begin{equation}
\label{eq:forward_closed_form}
S_k^{(\mathsf{out})}
=
\sqrt{\bar{\alpha}_k}S^{(\mathsf{out})}
+
\sqrt{1-\bar{\alpha}_k}\,\epsilon,
\end{equation}
where $\bar{\alpha}_k=\prod_{\kappa=1}^{k}\alpha_\kappa$ and $\epsilon\sim\mathcal{N}(0,I)$ is a random noise with the same dimensions as $S^{(\mathsf{out})}$, whose entries are independently drawn from a standard normal distribution with zero mean and unit variance. As $k$ increases, the contribution of the clean sequence decreases while that of the Gaussian noise increases; for large $k$, $S_k^{(\mathsf{out})}$ approaches isotropic Gaussian noise.

\noindent
$\bullet$ \textit{Context-conditioned denoiser and training objective.}
Given a noisy joint velocity sequence $S_k^{(\mathsf{out})}$ generated by the forward diffusion process, the denoiser $\epsilon_\theta$ is trained to predict the Gaussian noise added to the original sequence $S^{(\mathsf{out})}$. Specifically, the denoiser takes as input $S_k^{(\mathsf{out})}$, the corresponding diffusion level represented by $\beta_k$, and the context tokens $C^{(\mathsf{ctx})}$ obtained from Phase~1. We implement $\epsilon_\theta$ using a Transformer~\cite{yuan2021agentformer} that employs temporal self-attention to capture motion dependencies within each trajectory, agent-wise attention to capture interactions among robot and entity trajectories, and cross-attention to incorporate the scene and waypoint information encoded in $C^{(\mathsf{ctx})}$.
To train this denoiser across different noise levels, for each training sample, we randomly select a diffusion step $k$ and sample Gaussian noise $\epsilon\sim\mathcal{N}(0,I)$. We then construct the corresponding noisy sequence $S_k^{(\mathsf{out})}$ using~\eqref{eq:forward_diffusion} and train $\epsilon_\theta$ to predict the sampled noise $\epsilon$ by minimizing the following loss function:
\begin{equation}
\label{eq:diffusion_loss}
\mathcal{L}_{\mathsf{diff}}
= \mathbb{E}\left[
\left\|
\epsilon -
\epsilon_\theta\left(
S_k^{(\mathsf{out})},
\beta_k,
C^{(\mathsf{ctx})}
\right)
\right\|_2^2
\right].
\end{equation}
The expectation in~\eqref{eq:diffusion_loss} is taken over the training samples, randomly selected diffusion steps $k$, and sampled noise $\epsilon$. During training, this loss is minimized using gradient-based optimization, which updates the learnable parameters of both the denoiser $\epsilon_\theta$ and the learnable parameters of Phase~1 scene encoder that produces $C^{(\mathsf{ctx})}$. Consequently, the model jointly learns to encode the conditioning scene and denoise the coupled robot--entity future velocity sequences.

\noindent
$\bullet$ \textit{Reverse sampling process.}
At inference, the trained denoiser $\epsilon_\theta$ generates $N_{\mathrm{samp}}$ stochastic samples from the learned conditional joint trajectory distribution. For each sample, the reverse process starts from $S_K^{(\mathsf{out})}\sim\mathcal{N}(0,I)$ and proceeds from $k=K$ to $k=0$. At each diffusion level $k$, the denoiser predicts the noise contained in the current sequence $S_k^{(\mathsf{out})}$, which is then removed to obtain the corresponding clean joint velocity estimate $\hat S_k^{(\mathsf{out})}$. 
With a sampling stride $s$ and next diffusion level $k'=\max(k-s,0)$, the  reverse update is given by:
\begin{equation}
\label{eq:reverse_diffusion}
\begin{aligned}
\hat S_k^{(\mathsf{out})}
&=\frac{S_k^{(\mathsf{out})}-\sqrt{1-\bar\alpha_k}\,
\epsilon_\theta(S_k^{(\mathsf{out})},\beta_k,C^{(\mathsf{ctx})})}
{\sqrt{\bar\alpha_k}},\\
S_{k'}^{(\mathsf{out})}
&=\begin{cases}
\sqrt{\bar\alpha_{k'}}\,\hat S_k^{(\mathsf{out})}
+\sqrt{1-\bar\alpha_{k'}}\,z, & k'>0,\\
\hat S_k^{(\mathsf{out})}, & k'=0,
\end{cases}
\end{aligned}
\end{equation}
where $z\sim\mathcal{N}(0,I)$ is newly sampled Gaussian noise. Here, $\hat S_k^{(\mathsf{out})}$ represents the clean joint robot--entity velocity sequence estimated from the current noisy sequence $S_k^{(\mathsf{out})}$. If $k'>0$, a controlled amount of Gaussian noise, determined by $\bar{\alpha}_{k'}$, is added to this clean estimate to construct $S_{k'}^{(\mathsf{out})}$. We refer to this operation as \textit{re-noising}. Since $k'<k$, $S_{k'}^{(\mathsf{out})}$ corresponds to a lower diffusion level and is therefore less noisy than $S_k^{(\mathsf{out})}$; it then serves as the input to the next reverse step. This iterative process of estimating a clean sequence and re-noising it at progressively lower diffusion levels is repeated until $k'=0$, at which point no additional noise is added and the clean estimate becomes the final joint sample of robot plan velocities and entity-prediction velocities.
Hence, the reverse sampling process provides a practical mechanism for generating samples from the learned conditional distribution $p_\theta\!\left(S^{(\mathsf{out})}\mid C^{(\mathsf{ctx})}\right)$, corresponding to the sampling requirement in the objective of formulation $\bm{\mathcal{P}}$ in~\eqref{eq:MainPro}. However, the resulting samples are not guaranteed to satisfy constraints (C1)--(C5) as~\eqref{eq:reverse_diffusion} represents the \textit{unguided} reverse sampling process. To address this, Phase~3 incorporates safety-social guidance into the reverse sampling process by modifying only the robot component of each clean estimate $\hat S_k^{(\mathsf{out})}$ before re-noising.

\vspace{-5mm}
\subsection{Phase 3: Safety-Social Energy Guidance} 
\label{subsec:phase3}
Since Phase~2 samples are not guaranteed to satisfy constraints (C1)--(C5), Phase~3 guides the reverse sampling process toward constraint-compliant robot trajectories. 
To this end, following gradient-guided diffusion~\cite{dhariwal2021diffusion,jiang2023motiondiffuser}, we define a \textit{differentiable energy function} $\mathcal{J}$ over each clean joint velocity estimate (Sec.~\ref{Sec:EnergyFunctionDef}), where lower energy indicates greater compliance, and correct only the robot components before re-noising (see Appendix~\ref{app:aux_details}, Fig.~\ref{fig:app_guidance}); entity predictions remain unchanged.
 
 % We next define the four energy components and the corresponding guided correction.

% \begin{figure}[!t]
%     \centering
%     \includegraphics[width=\linewidth]{safety_social_guidance_phosphor}
%     \vspace{-7mm}
%     \caption{Illustration of one safety-social guidance step during reverse diffusion. Geometry-aware R2E/R2R collision margins, role-conditioned social regions, kinematic limits, and waypoint progress define the total energy. Its clipped gradient produces a bounded robot-only correction before re-noising to the next diffusion level; entity components are not directly modified by guidance.}
%     \label{fig:safety_social_guidance}
%     \vspace{-4mm}
% \end{figure}

\subsubsection{Differentiable energy function design} \label{Sec:EnergyFunctionDef}
We define $\mathcal{J}$ as a weighted differentiable energy function corresponding to constraints (C1)--(C5), evaluated on the integrated robot plans and predicted entity trajectories:
\begin{equation}
\label{eq:total_energy}
\hspace{-4mm}
\resizebox{0.45\textwidth}{!}{$
\begin{aligned}
\mathcal{J}(S^{(\mathsf{out})}){=}w_{\mathsf{col}}\mathcal{J}^{(\mathsf{col})}
+w_{\mathsf{soc}}\mathcal{J}^{(\mathsf{soc})}+w_{\mathsf{kin}}\mathcal{J}^{(\mathsf{kin})}
+w_{\mathsf{goal}}\mathcal{J}^{(\mathsf{goal})}.
\end{aligned}
$}\hspace{-4mm}
\end{equation}
The weights $w_{\mathsf{col}},w_{\mathsf{soc}},w_{\mathsf{kin}},w_{\mathsf{goal}}\geq 0$ control the importance of collision avoidance term $\mathcal{J}^{(\mathsf{col})}$, social compliance term $\mathcal{J}^{(\mathsf{soc})}$, kinematic feasibility term $\mathcal{J}^{(\mathsf{kin})}$, and task progress term $\mathcal{J}^{(\mathsf{goal})}$, respectively. These four terms are discussed next.

\textit{(i) Physical anti-collision potential energy ($\mathcal{J}^{(\mathsf{col})}$).}
To capture constraints (C1) and (C2), we define a differentiable potential that penalizes proximity between each robot--entity pair and each pair of robots. Specifically, for a robot $r_j$ and entity $o_i$, let
$d_{j,i}^t=\|\mathbf{p}_j^{(\mathsf{r},t)}-\mathbf{p}_i^{(\mathsf{o},t)}\|_2$
denote their distance at time $t$, and for robots $r_j$ and $r_{j'}$, let
$d_{j,j'}^t=\|\mathbf{p}_j^{(\mathsf{r},t)}-\mathbf{p}_{j'}^{(\mathsf{r},t)}\|_2$.
We define the collision potential
$\phi_{\mathsf{col}}(d;m)=\exp[-d^2/(2\sigma_{\mathsf{col}}^2)]+\mathrm{ReLU}(m-d)^2$,
which provides smooth repulsion as the pairwise distance $d$ decreases and imposes an additional penalty when $d$ falls below the safety margin $m$. The resulting collision energy is then defined as follows:
\begin{equation}
\hspace{-3mm}
\resizebox{0.45\textwidth}{!}{$
\begin{aligned}
\mathcal{J}^{(\mathsf{col})}
{=}\hspace{-.8mm}\sum_{t=1}^{T^{(\mathsf{prd})}}\hspace{-.7mm}\Bigg[\hspace{-.3mm}
\sum_{r_j\in\mathcal R}\hspace{-.3mm}\Big[\hspace{-.99mm}\sum_{o_i\in\mathcal O_{\mathcal N}^{0}}
\hspace{-1.2mm}\phi_{\mathsf{col}}(d_{j,i}^t;m_{j,i}){+}w_{\mathsf{R2R}}\sum_{j<j'}
\phi_{\mathsf{col}}(d_{j,j'}^t;m_{j,j'})\Big]\hspace{-.3mm}\Bigg]\hspace{-.5mm},
\end{aligned}
$}
\hspace{-3mm}
\end{equation}
where $m_{j,i}$ and $m_{j,j'}$ are the safety margins defined in (C1) and (C2), and $w_{\mathsf{R2R}}\geq 0$ controls the relative importance of robot-to-robot collision avoidance (the summation index $j<j'$ ensures that each distinct robot pair is counted only once).

\textit{(ii) Heterogeneous anisotropic social potential ($\mathcal{J}^{(\mathsf{soc})}$).}
To capture the role-conditioned social compliance constraint in (C4), we define a differentiable potential that penalizes robots for entering the social comfort regions of neighboring entities in their forward half-plane. Specifically, for each robot $r_j$ and entity $o_i$, we define the normalized relative position
$\boldsymbol{\delta}_{i,j}^t=(d_{i,j}^{(\mathsf{lon},t)}/\sigma_i^{\mathsf{lon}},d_{i,j}^{(\mathsf{lat},t)}/\sigma_i^{\mathsf{lat}})$
where
$\sigma_i^{\mathsf{lon}}=\sigma^{\mathsf{lon}}(\mathcal{C}_i^{(\mathsf{o,role})})$
and
$\sigma_i^{\mathsf{lat}}=\sigma^{\mathsf{lat}}(\mathcal{C}_i^{(\mathsf{o,role})})$
are defined in (C4). The resulting social compliance energy is then defined as:
\begin{equation}
\hspace{-3mm}
\resizebox{0.40\textwidth}{!}{$
\begin{aligned}
\mathcal{J}^{(\mathsf{soc})}
=\hspace{-.2mm}\sum_{t=1}^{T^{(\mathsf{prd})}}
\sum_{r_j\in\mathcal{R}}
\sum_{o_i\in\mathcal O_{\mathcal N}^{0}}
\mathbb{I}\!\left[d_{i,j}^{(\mathsf{lon},t)}\geq0\right]
\exp\!\left(-\|\boldsymbol{\delta}_{i,j}^t\|_2^2\right), 
\end{aligned}
$}
\end{equation}
where the term
$\exp(-\|\boldsymbol{\delta}_{i,j}^t\|_2^2)$
assigns a larger penalty when the robot is closer to the entity within its normalized social space, while
$\mathbb{I}[d_{i,j}^{(\mathsf{lon},t)}\geq0]$
restricts this penalty to entities located in the robot's forward half-plane.

\textit{(iii) Kinematic potential energy ($\mathcal{J}^{(\mathsf{kin})}$).}
To capture the kinematic feasibility constraint in (C3), we penalize violations of the prescribed translational-speed, lateral-velocity, and angular-velocity limits. Specifically, each penalty remains zero when the corresponding kinematic quantity is within its allowable range and increases quadratically once the limit is exceeded. The resulting kinematic energy is defined as follows:
\begin{equation}
\hspace{-3mm}
\resizebox{0.455\textwidth}{!}{$
\begin{aligned}
\mathcal{J}^{(\mathsf{kin})}
{=} \hspace{-1.1mm}\sum_{t=1}^{T^{(\mathsf{prd})}}\hspace{-0.85mm}\sum_{r_j\in\mathcal R}\hspace{-0.7mm}\Big[\hspace{-0.4mm}
\big[\|\mathbf v_j^t\|_2-v_{\max}\big]_+^2{+}\big[|v_j^{(\mathsf{lat},t)}|{-}v_{\max}^{\mathsf{lat}}\big]_+^2
{+}\big[|\omega_j^t|-\omega_{\max}\big]_+^2\Big]\hspace{-0.2mm},
\end{aligned}
$}\hspace{-3mm}
\end{equation}
where $[x]_+=\mathrm{ReLU}(x)=\max(x,0)$. 

\noindent
\textit{(iv) Task-progress potential energy ($\mathcal{J}^{(\mathsf{goal})}$).}
To capture the task-progress requirement in (C5), we penalize the terminal distance between each robot and its assigned waypoint. Specifically, the task-progress energy is defined as follows:
\begin{equation}
\label{eq:goal_energy}
\resizebox{0.24\textwidth}{!}{$
\begin{aligned}
\mathcal{J}^{(\mathsf{goal})}
=
\sum_{r_j\in\mathcal{R}}
\left\|
\mathbf{p}_j^{(\mathsf{r},T^{(\mathsf{prd})})}
-
\mathbf{g}_j
\right\|_2^2 .
\end{aligned}
$}
\end{equation}

\subsubsection{Guided correction algorithm design}\label{Sec:Correct}
We now integrate the energy-guided correction into the reverse sampling process of Phase~2. Recall that, in the unguided reverse update in~\eqref{eq:reverse_diffusion}, the denoiser first produces the clean joint velocity estimate $\hat S_k^{(\mathsf{out})}$, which is then directly re-noised to obtain $S_{k'}^{(\mathsf{out})}$. Phase~3 modifies this process by inserting an energy-guided correction between these two operations: $\hat S_k^{(\mathsf{out})}$ is first corrected to obtain $\widetilde S_k^{(\mathsf{out})}$, and this guided estimate, rather than $\hat S_k^{(\mathsf{out})}$, is subsequently re-noised to continue the reverse sampling process. Specifically, we use the energy function $\mathcal{J}$ defined in~\eqref{eq:total_energy} to guide the clean velocity estimates generated during the reverse sampling process toward greater constraint compliance. Specifically, at diffusion level $k$, SAGE first evaluates $\mathcal{J}(\hat S_k^{(\mathsf{out})})$ on the clean joint robot--entity velocity estimate $\hat S_k^{(\mathsf{out})}$ obtained from~\eqref{eq:reverse_diffusion}. The corresponding gradient
$\nabla_{\hat S_k^{(\mathsf{out})}}\mathcal{J}(\hat S_k^{(\mathsf{out})})$
indicates how changes to the estimated velocities affect the total energy. Since only robot trajectories are controllable, we apply a mask $\mathcal{M}_{\mathcal R}$ that retains the gradient components corresponding to robot velocities and sets the entity components to zero, and then modify the robot velocities in the negative-gradient direction to reduce $\mathcal{J}$. Specifically, the guided correction and subsequent re-noising are performed as follows:
\begin{equation}
\label{eq:guided_correction}
\begin{aligned}
\mathbf g_k
&=\mathrm{Clip}_{\mathsf{grad}}\!\left(
\mathcal M_{\mathcal R}\nabla_{\hat S_k^{(\mathsf{out})}}
\mathcal J(\hat S_k^{(\mathsf{out})})\right),\\
\widetilde S_k^{(\mathsf{out})}
&=\Pi_{\mathsf{vel}}\!\left(
\hat S_k^{(\mathsf{out})}
-\mathrm{Clip}_{\mathsf{step}}(\lambda\mathbf g_k)
\right),\\
S_{k'}^{(\mathsf{out})}
&=\begin{cases}
\sqrt{\bar\alpha_{k'}}\,\widetilde S_k^{(\mathsf{out})}
+\sqrt{1-\bar\alpha_{k'}}\,z, & k'>0,\\
\widetilde S_k^{(\mathsf{out})}, & k'=0.
\end{cases}
\end{aligned}
\end{equation}
 In \eqref{eq:guided_correction}, the operator $\mathrm{Clip}_{\mathsf{grad}}(\cdot)$ applies trajectory-wise norm clipping to the robot-only gradient $\mathcal M_{\mathcal R}\nabla_{\hat S_k^{(\mathsf{out})}}
\mathcal J(\hat S_k^{(\mathsf{out})})$, resulting in the bounded gradient $\mathbf g_k$. The parameter $\lambda$ controls the correction strength, while $\mathrm{Clip}_{\mathsf{step}}$ bounds the resulting correction step. Also, the operator $\Pi_{\mathsf{vel}}(\cdot)$ caps the corrected robot speeds at $\gamma_{\mathsf{vel}}v_{\max}$, with $\gamma_{\mathsf{vel}}=1.25$, while leaving the entity velocities unchanged, resulting in the guided clean joint velocity estimate $\widetilde S_k^{(\mathsf{out})}$. Finally, $z\sim\mathcal{N}(0,I)$ denotes the Gaussian noise used to re-noise $\widetilde S_k^{(\mathsf{out})}$ when $k'>0$.
In words, at each reverse sampling step, SAGE computes how the total energy changes with the clean joint velocity estimate, retains only the gradient corresponding to the controllable robots, and adjusts their velocities in the negative-gradient direction to reduce the energy. The resulting correction and robot speeds are bounded to prevent excessive changes, while the predicted entity velocities remain unchanged. The corrected estimate is then re-noised to continue the reverse sampling process when $k'>0$; when $k'=0$, it becomes the final joint velocity sample.

%\subsection{Framework Coordination}
%By integrating these three phases, the framework proposed in this paper provides a structured sampling solution for the constrained planning problem $\mathcal{P}$. The HGT encoding in Phase 1 injects statistical priors of heterogeneous interactions and task-conditioned robot intent into the generative model, ensuring that the joint trajectory distribution generated in Phase 2 aligns macroscopically with the social game logic among entities of different categories, thereby reducing the magnitude of corrections required for safety-social energy guidance in Phase 3; The differentiable potential field in Phase 3 provides explicit constraint-aware corrections for long-tail extreme scenarios that Phase 2 cannot cover, encouraging the final planned trajectory to better satisfy the requirements of task progress, physical safety, and social compliance. The three phases form a complementary mechanism of ``implicit prior proposal (Phase 1 + Phase 2) -- explicit rule fallback (Phase 3)'': Relying solely on Phases 1 and 2 lacks strict constraints when facing out-of-distribution scenarios; relying solely on Phase 3's safety-social energy guidance without $C^{(\mathsf{ctx})}$ contextual conditions causes the guidance gradient to conflict with the diffusion model's unconditional denoising direction, triggering trajectory oscillations. The synergy between the two enables the framework to simultaneously possess statistical naturalness and fluidity alongside robust compliance at the constraint level.

\vspace{-2mm}
\section{Experiments}
\label{sec:experiments}

In the following, we numerically evaluate SAGE
in terms of task effectiveness, safety, social compliance, guidance controllability, and R2R coordination.

%We begin by verifying that the framework generates safe and socially-compliant trajectories on real-world data (Section~\ref{subsec:main_results}). Having established effectiveness, we analyze whether the guidance mechanism is controllable and characterize its safety--accuracy trade-off (Section~\ref{subsec:guidance}). We then report selected architecture diagnostics to clarify the role of key modules (Section~\ref{subsec:ablation}). Finally, we stress-test the system under explicit multi-robot coordination scenarios that recorded datasets cannot provide (Section~\ref{subsec:scalability}). Qualitative examples and a structured discussion conclude the evaluation (Sections~\ref{subsec:qualitative}--\ref{subsec:discussion}).

\vspace{-2mm}
\subsection{Experimental Setup}
\label{subsec:setup}
We consider two evaluation regimes: real-world trajectory data for safety and social compliance, and controlled heterogeneous simulations for task performance and scalability.

\noindent
$\bullet$ \textit{Datasets and rationale.}
We employ three data sources, each serving a distinct purpose. \textit{(i) ETH/UCY}~\cite{alahi2016social} contains five pedestrian scenes (ETH, HOTEL, UNIV, ZARA1, and ZARA2). Since this dataset lacks semantic labels and robot waypoints, we treat a target pedestrian as the robot and its future endpoint as the waypoint. \textit{(ii) SDD} provides aerial trajectories with official object annotations~\cite{robicquet2016learning}. We map Pedestrians to PHAs, Bikers/Skaters/Carts to HNAAs, and Cars/Buses to SSAAs, making SDD the primary source for role-aware social-compliance evaluation. \textit{(iii) Controlled heterogeneous simulations} complement recorded data by varying robot population, entity density, role composition, and start-waypoint conflicts. 
Robots receive independently assigned waypoints across open, crossing, corridor, bottleneck, and intersection layouts, enabling task and scalability evaluation in Sec.~\ref{subsec:scalability}.

% Robots receive commanded waypoints independent of their future trajectories across open, crossing, corridor, bottleneck, and intersection layouts, enabling genuine task and scalability evaluation in Sec.~\ref{subsec:scalability}.

\noindent
$\bullet$ \textit{Evaluation protocol.}
% We set $T^{(\mathsf{obs})}=8$, $T^{(\mathsf{prd})}=12$, and $\Delta t=0.4\,\mathrm{s}$. Unless noted, evaluation uses $N_{\mathrm{samp}}=20$ stochastic samples with a reverse-diffusion step stride of 10. SAGE w/ and SAGE w/o guidance share the same checkpoint and differ only in whether the safety-social gradient is applied; w/ and w/o denote these settings.
We set $T^{(\mathsf{obs})}=8$, $T^{(\mathsf{prd})}=12$, and $\Delta t=0.4\mathrm{s}$. Unless noted, we use $N_{\mathrm{samp}}=20$ stochastic samples and a reverse-diffusion stride of 10. 

% SAGE w/ and w/o guidance methods  differ only in the inclusion of safety-social guidance component of SAGE discussed in Sec.~\ref{subsec:phase3}.

\noindent
$\bullet$ \textit{Safety and social-compliance metrics.}
For \textit{physical safety}, collision rate (CR) metric measures the fraction of robot-entity and robot-robot interactions across the prediction horizon with separation below the collision margin, while minimum separation (MD) metric measures the minimum separation over the prediction horizon. For \textit{social compliance}, personal-space intrusion rate (PIR) metric measures the fraction of robot--entity interactions across the prediction horizon within the personal-space margin, while social-violation rate (SVR) metric measures the fraction within the role-conditioned ellipse in~\eqref{eq:outside-SC}. Speed-violation rate (Sp-VR) metric measures the fraction of robot states exceeding the maximum speed, and Energy measures the total guidance objective $\mathcal J$ in~\eqref{eq:total_energy}, averaged over evaluation scenes and generated trajectory samples.

% \noindent
% $\bullet$ \textit{Task/accuracy metrics and aggregation.}
% For \textit{task progress}, Goal-FDE metric measures the terminal distance to the assigned waypoint, and Goal-SR measures the fraction of active robots with assigned waypoints whose terminal waypoint distance does not exceed $d^{(\mathsf{goal})}$. Also, ADE/FDE metric measures mean/final Euclidean errors against recorded trajectories. In metric names, the prefixes R- and E- denote robot and entity quantities, respectively (e.g., R-ADE and E-ADE). ADE/FDE and goal metrics report the best-performing candidate among 20 samples under each metric, whereas safety-social metrics report the mean over all 20 generated candidates; no downstream policy for selecting an executed trajectory is evaluated. Lower values are better except for MD and Goal-SR. The prefixes R2E- and R2R- denote robot-to-entity and robot-to-robot interactions, respectively, and are distinct from the R2N/N2R graph relations in Phase~1. Detailed preprocessing, comfort margins, thresholds, and model settings are provided in Appendix~\ref{sec:supp_settings}.

\noindent
$\bullet$ \textit{Task/accuracy metrics.}
For \textit{task progress}, Goal-FDE measures the terminal distance to the assigned waypoint, while Goal-SR measures the fraction of robots reaching within $d^{(\mathsf{goal})}$ of their assigned waypoints. ADE/FDE measure the mean/final Euclidean errors against recorded trajectories. The prefixes R- and E- denote robot and entity quantities, respectively (e.g., R-ADE and E-ADE). ADE/FDE and goal metrics report the best of 20 trajectory samples, while safety-social metrics are averaged over all 20 samples. Lower values are better except for MD and Goal-SR. Details of preprocessing, comfort margins, thresholds, and model settings are provided in Appendix~\ref{sec:supp_settings}.

\begin{table}[!t]
\vspace{-8.5mm}
\centering
\caption{Results of real-world scenes. ETH/UCY values are averaged over five scenes; SDD uses official semantic annotations.}
\label{tab:main_results}
\vspace{-2mm}
\scriptsize
\resizebox{\columnwidth}{!}{
\setlength{\tabcolsep}{1.2pt}
\begin{tabular}{llccccccc}
\hline
\rowcolor{gray!25}
\textbf{Dataset} & \textbf{Method} & \textbf{R-ADE$\downarrow$} & \textbf{Goal-FDE$\downarrow$} & \textbf{CR$\downarrow$} & \textbf{MD$\uparrow$} & \textbf{PIR$\downarrow$} & \textbf{SVR$\downarrow$} & \textbf{Energy$\downarrow$} \\
\hline
\multirow{5}{*}{ETH/UCY} & Constant Velocity & 0.4760 & 1.0092 & 0.0471 & 0.0398 & 0.3407 & 0.1456 & 20.4553 \\
& Trajectron++ & 0.1461 & 0.0968 & 0.0416 & 0.0469 & 0.3440 & 0.1355 & 20.0778 \\
& MID & 0.1163 & 0.0929 & 0.0409 & 0.0510 & 0.3547 & 0.1482 & 20.1621 \\
& SAGE w/o guidance & 0.1152 & 0.0937 & 0.0441 & 0.0471 & 0.3560 & 0.1479 & 20.3761 \\
& SAGE w/ guidance & 0.1174 & 0.0947 & 0.0353 & 0.0623 & 0.3457 & 0.1248 & 19.1257 \\
\hline
\multirow{4}{*}{SDD} & Constant Velocity & 0.7290 & 1.6818 & 0.1701 & 0.0572 & 0.1560 & 0.0892 & 28.7860 \\
& Trajectron++ & 0.1391 & 0.1188 & 0.1798 & 0.0875 & 0.1645 & 0.0965 & 27.6866 \\
& SAGE w/o guidance & 0.1151 & 0.0700 & 0.1763 & 0.0582 & 0.1612 & 0.0925 & 26.1494 \\
& SAGE w/ guidance  & 0.1241 & 0.0813 & 0.1729 & 0.0694 & 0.1576 & 0.0679 & 24.4608 \\
\hline
\end{tabular}}
\vspace{-7mm}
\end{table}

% \noindent
% $\bullet$ \textit{Comparative benchmarks.}
% On ETH/UCY dataset, we compare against three reference methods under the same proxy-navigation protocol: \textit{Constant Velocity} (deterministic short-horizon extrapolation), \textit{Trajectron++}~\cite{salzmann2020trajectron++} (CVAE-based stochastic prediction), and \textit{MID}~\cite{gu2022stochastic} (diffusion-based prediction without guidance). On SDD, we use Constant Velocity and Trajectron++. We also report the complete framework (SAGE) and an ablation that disables its Phase~3 (\textit{SAGE w/o guidance}).

$\bullet$ \textit{Benchmarks.}
On ETH/UCY dataset, we compare against three methods under the same proxy-navigation protocol: \textit{Constant Velocity}, \textit{Trajectron++}~\cite{salzmann2020trajectron++}, and \textit{MID}~\cite{gu2022stochastic}. On SDD, we compare against Constant Velocity and Trajectron++. We also evaluate SAGE with and without Phase~3 guidance discussed in Sec.~\ref{subsec:phase3} (\textit{SAGE w/ guidance} and \textit{SAGE w/o guidance}). All simulations use an NVIDIA RTX 4090 GPU.

\vspace{-2mm}
\subsection{Safety/Social Compliance on ETH/UCY and SDD}
\label{subsec:main_results}

% We first evaluate the  SAGE framework on real trajectory data and isolate the contribution of its inference-time safety-social guidance by comparing it with SAGE w/o guidance. Table~\ref{tab:main_results} reports summarized results across ETH/UCY (five-scene average) and SDD, where two findings emerge. 
% \textit{First}, guidance produces consistent safety improvements. On ETH/UCY, SAGE reduces CR by 20.0\% (0.0441$\rightarrow$0.0353), SVR by 15.6\%, and Energy by 6.1\%, while increasing MD by 1.32$\times$ relative to SAGE w/o guidance. Notably, SAGE w/o guidance is slightly \emph{less} safe than Trajectron++ and MID on several metrics (e.g., CR 0.0441 vs.\ 0.0416 and 0.0409), confirming that the improvement comes from guidance rather than the diffusion prior alone. On SDD, guidance reduces SVR by 26.6\% and Energy by 6.5\%, while the CR improvement is smaller (1.9\%, 0.1763$\rightarrow$0.1729). \textit{Second}, the trade-off between safety and accuracy is modest: SAGE w/o guidance is competitive with the best learned baseline on ETH/UCY (R-ADE 0.1152 vs.\ MID 0.1163 and Trajectron++ 0.1461), and guidance increases R-ADE by only 1.9\% (0.1152$\rightarrow$0.1174). Commanded-waypoint evaluation is presented in Sec.~\ref{subsec:scalability}, and complete per-scene and role-wise results are reported in Appendix~\ref{sec:supp_recorded_diagnostics}.
We first evaluate SAGE on real-world trajectory data, with Table~\ref{tab:main_results} summarizing the results on ETH/UCY and SDD. Two findings emerge from comparing SAGE with and without guidance.
\textit{First}, compared with SAGE w/o guidance, SAGE consistently improves safety and social compliance. On ETH/UCY, SAGE reduces CR by 20.0\% (0.0441$\rightarrow$0.0353), SVR by 15.6\%, and Energy by 6.1\%, while increasing MD by 1.32$\times$ when compared with SAGE w/o guidance. On SDD, SAGE reduces SVR by 26.6\% and Energy by 6.5\%, with a smaller reduction in CR (1.9\%) when compared with SAGE w/o guidance.
\textit{Second}, these improvements come with a modest accuracy trade-off compared with SAGE w/o guidance. On ETH/UCY, SAGE w/o guidance achieves an R-ADE of 0.1152, compared with 0.1163 for MID and 0.1461 for Trajectron++, while SAGE achieves an R-ADE of 0.1174, corresponding to only a 1.9\% increase. Complete per-scene and role-wise results are provided in Appendix~\ref{sec:supp_recorded_diagnostics}.

\vspace{-4mm}
\subsection{Guidance Mechanism: Safety--Accuracy Trade-off}
\label{subsec:guidance}
% The above results treat guidance as a binary switch. We now sweep the guidance scale $\lambda$ in~\eqref{eq:guided_correction}, which controls the strength of the safety-social energy gradient during reverse diffusion, to answer two questions: \textit{(i)} is the correction strength controllable, and \textit{(ii)} what accuracy cost accompanies each safety level?
% As $\lambda$ increases from 0 to 0.05, Fig.~\ref{fig:guidance_curves_main} shows that CR decreases by 20.0\%, SVR by 15.6\%, and Energy by 6.1\%, while R-ADE increases by only 1.9\%.  The modest degradation indicates that the learned prior already proposes plausible trajectories and guidance acts as a lightweight correction layer. Unless otherwise specified, subsequent experiments use $\lambda=0.05$. Thus, $\lambda$ provides inference-time control over the safety--accuracy trade-off without retraining.

\begin{figure}[!t]
\vspace{-2.5mm}
\centering
\subfigure[]{\includegraphics[width=0.485\columnwidth, trim = 0 11 0 14, clip]{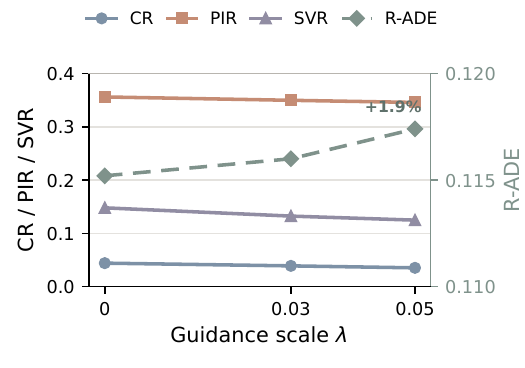}\label{fig:guidance_curves_main}}
\hfill
\subfigure[]{\includegraphics[width=0.485\columnwidth, trim = 0 11 0 10, clip]{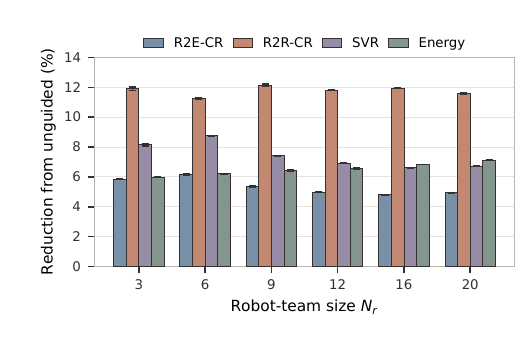}\label{fig:scalability_curves}}
\vspace{-2mm}
\caption{Guidance effects across evaluation regimes. (a) Safety, social compliance, and trajectory accuracy versus guidance scale $\lambda$ on ETH/UCY (five-scene average); CR, PIR, and SVR use the left axis, while R-ADE uses the right axis. (b) Relative safety-social improvement in the multi-robot scalability sweep; bars report reductions over unguided sampling averaged over three seeds, and error bars denote standard deviation.}
\label{fig:guidance_summary}
\vspace{-5mm}
\end{figure}

We next evaluate the effect of the guidance scale $\lambda$, which controls the strength of the safety-social energy gradient applied during reverse diffusion dictated by~\eqref{eq:guided_correction}. As shown in Fig.~\ref{fig:guidance_curves_main}, increasing $\lambda$ from 0 to 0.05 reduces CR and SVR by 20.0\% and 15.6\%, respectively, while R-ADE increases by only 1.9\% (0.1152$\rightarrow$0.1174). Thus, $\lambda$ provides inference-time control over the safety--accuracy trade-off without retraining. Unless specified, we use $\lambda=0.05$ in experiments.
\begin{table}[!t]
\vspace{-8.5mm}
\centering
\caption{Multi-robot scalability ($N_e=50$, $N_{\mathrm{samp}}=20$). Metric values are averaged over three seeds, each with 1,024 scenes.}
\label{tab:scalability}
\vspace{-2mm}
\scriptsize
\resizebox{\columnwidth}{!}{
\setlength{\tabcolsep}{2.0pt}
\begin{tabular}{cccccccccc}
\hline
\rowcolor{gray!25} \textbf{$N_r$} & \textbf{Guidance} & \textbf{Goal-FDE}$\downarrow$ & \textbf{R2E-CR}$\downarrow$ & \textbf{R2R-CR}$\downarrow$ & \textbf{R2E-MD}$\uparrow$ & \textbf{R2R-MD}$\uparrow$ & \textbf{SVR}$\downarrow$ & \textbf{Energy}$\downarrow$ & \textbf{Runtime} \\
\hline
3 & w/o & 5.872 & 0.0418 & 0.0851 & 0.0206 & 0.0578 & 0.0216 & 292.2 & 85.0 ms \\
3 & w/ & 5.928 & 0.0394 & 0.0750 & 0.0241 & 0.0838 & 0.0198 & 274.7 & 92.1 ms \\
6 & w/o & 5.813 & 0.0372 & 0.0781 & 0.0136 & 0.0308 & 0.0185 & 594.9 & 90.5 ms \\
6 & w/  & 5.868 & 0.0349 & 0.0693 & 0.0171 & 0.0465 & 0.0169 & 557.9 & 99.6 ms \\
9 & w/o & 5.430 & 0.0397 & 0.0924 & 0.0122 & 0.0194 & 0.0202 & 966.3 & 96.6 ms \\
9 & w/ & 5.465 & 0.0375 & 0.0812 & 0.0146 & 0.0286 & 0.0187 & 904.3 & 106.1 ms \\
12 & w/o & 5.420 & 0.0399 & 0.0951 & 0.0106 & 0.0138 & 0.0205 & 1369.8 & 103.1 ms \\
12 & w/ & 5.448 & 0.0379 & 0.0839 & 0.0117 & 0.0207 & 0.0191 & 1279.9 & 113.8 ms \\
16 & w/o & 5.319 & 0.0383 & 0.0910 & 0.0090 & 0.0116 & 0.0193 & 1881.1 & 123.0 ms \\
16 & w/  & 5.330 & 0.0365 & 0.0801 & 0.0111 & 0.0165 & 0.0180 & 1752.6 & 137.0 ms \\
20 & w/o & 5.432 & 0.0387 & 0.0869 & 0.0078 & 0.0094 & 0.0196 & 2473.6 & 132.0 ms \\
20 & w/  & 5.444 & 0.0368 & 0.0769 & 0.0093 & 0.0141 & 0.0183 & 2297.5 & 147.2 ms \\
\hline
\end{tabular}}
\vspace{-1mm}
\end{table}

\vspace{-4mm}
\subsection{Multi-Robot Evaluation: Coordination and Scalability}
\label{subsec:scalability}
We use controlled simulations to evaluate task performance and scalability as the robot team grows. We evaluate \textit{(i)} the effectiveness of guidance with increasing team size and \textit{(ii)} the individual contributions of R2R graph edges and R2R collision guidance to safety and waypoint progress.

\noindent
$\bullet$ \textit{Scalability of guided sampling.}
% Table~\ref{tab:scalability} sweeps $N_r=|\mathcal{R}|=3$ to $20$ with $N_e=|\mathcal{O}|=50$, yielding three findings. \textit{First}, guidance improves all reported safety-social metrics at every team size (Fig.~\ref{fig:scalability_curves}). Averaged over all $N_r$, it reduces R2E-CR by 5.35\%, R2R-CR by 11.75\%, SVR by 7.43\%, and Energy by 6.52\%, while increasing R2E-MD by 1.19$\times$ and R2R-MD by 1.48$\times$. Because CR is normalized over active pair-time instances, it need not increase monotonically with $N_r$. \textit{Second}, gains are larger for the more challenging R2R interactions because unguided trajectories already maintain substantial R2E separation. At $N_r=20$, R2R-CR still falls from 0.0869 to 0.0769 and R2R-MD rises from 0.0094 to 0.0141. \textit{Third}, task and computational costs remain modest: Goal-FDE changes by less than 2\% across all scales, and guided runtime rises from 92.1\,ms at $N_r=3$ to 147.2\,ms at $N_r=20$, below the 0.4-s sampling interval and supporting online replanning at all evaluated scales.
Table~\ref{tab:scalability} evaluates $N_r=|\mathcal{R}|=3$ to $20$ with $N_e=|\mathcal{O}|=50$. \textit{First}, guidance improves safety and social compliance-related metrics across all team sizes (further visualized in Fig.~\ref{fig:scalability_curves}): averaged over all $N_r$, it reduces R2E-CR by 5.35\%, R2R-CR by 11.75\%, SVR by 7.43\%, and Energy by 6.52\%, while increasing R2E-MD by 1.19$\times$ and R2R-MD by 1.48$\times$. \textit{Second}, the improvement is particularly pronounced for R2R safety: at $N_r=20$, guidance reduces R2R-CR from 0.0869 to 0.0769 and increases R2R-MD from 0.0094 to 0.0141. \textit{Third}, these improvements have modest task and computational costs: Goal-FDE changes by less than 2\% across all team sizes, while guided runtime increases from 92.1\,ms at $N_r=3$ to 147.2\,ms at $N_r=20$, remaining below the 0.4s sampling interval.

% \noindent
% $\bullet$ \textit{Roles of the two R2R mechanisms.}
% Table~\ref{tab:r2r_ablation} ablates the R2R graph edges in Phase~1 and the R2R collision-guidance term in Phase~3. The four variants enable neither component, only the graph edges, only the R2R collision-guidance term, or both components (complete SAGE); all other components remain unchanged. Relative to the variant with neither component, R2R guidance alone reduces R2R-CR by 10.4\% at $N_r=12$ (0.0892$\rightarrow$0.0799) and 11.0\% at $N_r=20$ (0.0858$\rightarrow$0.0764), while increasing R2R-MD by 1.47$\times$ at both scales. Thus, at the two evaluated team sizes, R2R collision guidance provides the larger direct safety gain. Complete SAGE obtains the best Goal-FDE in both settings, showing that graph edges complement collision guidance by preserving waypoint progress.

\noindent
$\bullet$ \textit{Roles of the two R2R mechanisms in SAGE.}
Table~\ref{tab:r2r_ablation} evaluates the individual contributions of R2R graph edges in Phase~1 and R2R collision guidance in Phase~3 of SAGE. Compared with using neither component, R2R collision guidance alone reduces R2R-CR by 10.4\% at $N_r=12$ and 11.0\% at $N_r=20$, while increasing R2R-MD by 1.47$\times$ at both team sizes. Complete SAGE achieves the best (i.e., the lowest) Goal-FDE in both settings, indicating that R2R graph edges complement collision guidance in maintaining waypoint progress.

\begin{table}[!t]
\vspace{-8.5mm}
\centering
\caption{Ablation of R2R graph edges and collision guidance.}
\label{tab:r2r_ablation}
\vspace{-2mm}
\scriptsize
\setlength{\tabcolsep}{1.8pt}
\begin{tabular}{clccc}
\hline
\rowcolor{gray!25} 
$N_r$ & \textbf{Variant} & \textbf{Goal-FDE}$\downarrow$ & \textbf{R2R-CR}$\downarrow$ & \textbf{R2R-MD}$\uparrow$ \\
\hline
12 & Neither component & 5.574 & 0.0892 & 0.0144 \\
12 & Graph edges only & 5.488 & 0.0934 & 0.0146 \\
12 & R2R guidance only & 5.534 & \textbf{0.0799} & \textbf{0.0212} \\
12 & Both (SAGE) & \textbf{5.448} & 0.0839 & 0.0207 \\
\hline
20 & Neither component & 5.557 & 0.0858 & 0.0095 \\
20 & Graph edges only & 5.510 & 0.0859 & 0.0094 \\
20 & R2R guidance only & 5.486 & \textbf{0.0764} & 0.0140 \\
20 & Both (SAGE) & \textbf{5.444} & 0.0769 & \textbf{0.0141} \\
\hline
\end{tabular}
\vspace{-1mm}
\end{table}

\begin{figure}[!t]
\vspace{-3mm}
\centering
\includegraphics[width=\columnwidth, trim= 10 5 10 40, clip]{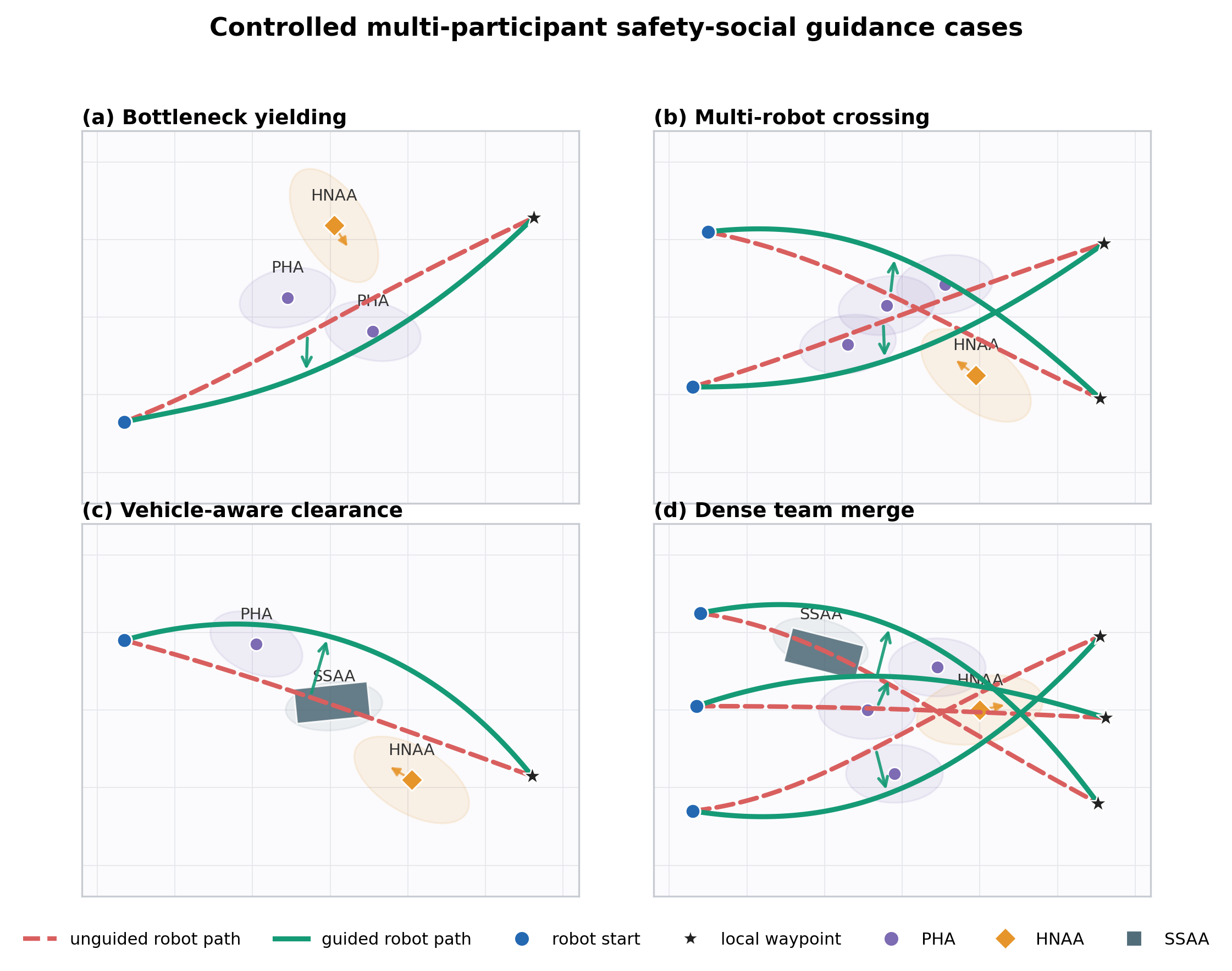}
\vspace{-8mm}
\caption{Controlled cases: bottleneck yielding, multi-robot crossing, vehicle-aware clearance, and dense merging. Dashed red: unguided; solid green: guided; blue circles: robot starts.}
\label{fig:synthetic_qualitative}\vspace{-5mm}
\end{figure}

% \vspace{-2mm}
% \subsection{Qualitative Analysis}
% \label{subsec:qualitative}
% Fig.~\ref{fig:synthetic_qualitative} visualizes representative multi-robot cases in controlled heterogeneous simulations. In each case, the robot team, waypoints, heterogeneous entities, and scene layout are fixed; only the guidance switch changes. We observe three patterns. \textit{First}, guidance corrections are localized: unguided trajectories that already avoid entities remain largely unchanged, while those encroaching on social fields or collision boundaries receive targeted repulsion. \textit{Second}, the correction magnitude adapts to entity type: robots maintain larger clearance from PHAs than from SSAAs, consistent with the role-conditioned social field design. \textit{Third}, guidance preserves waypoint progress: in all four panels, guided trajectories terminate near the same waypoints as unguided ones, illustrating how $w_{\mathsf{goal}}$ balances safety against task completion. Fig.~\ref{fig:sdd_qualitative} further shows that, in recorded SDD scenes, guidance locally redirects robot-proxy trajectories away from nearby role-labeled entities while leaving unaffected trajectory segments close to their unguided counterparts.

\vspace{-5mm}
\subsection{Qualitative Analysis}
\label{subsec:qualitative}
\begin{figure}[!t]
\vspace{-7mm}
\centering
\includegraphics[width=0.15\textwidth, trim=17 10 30 5, clip]
{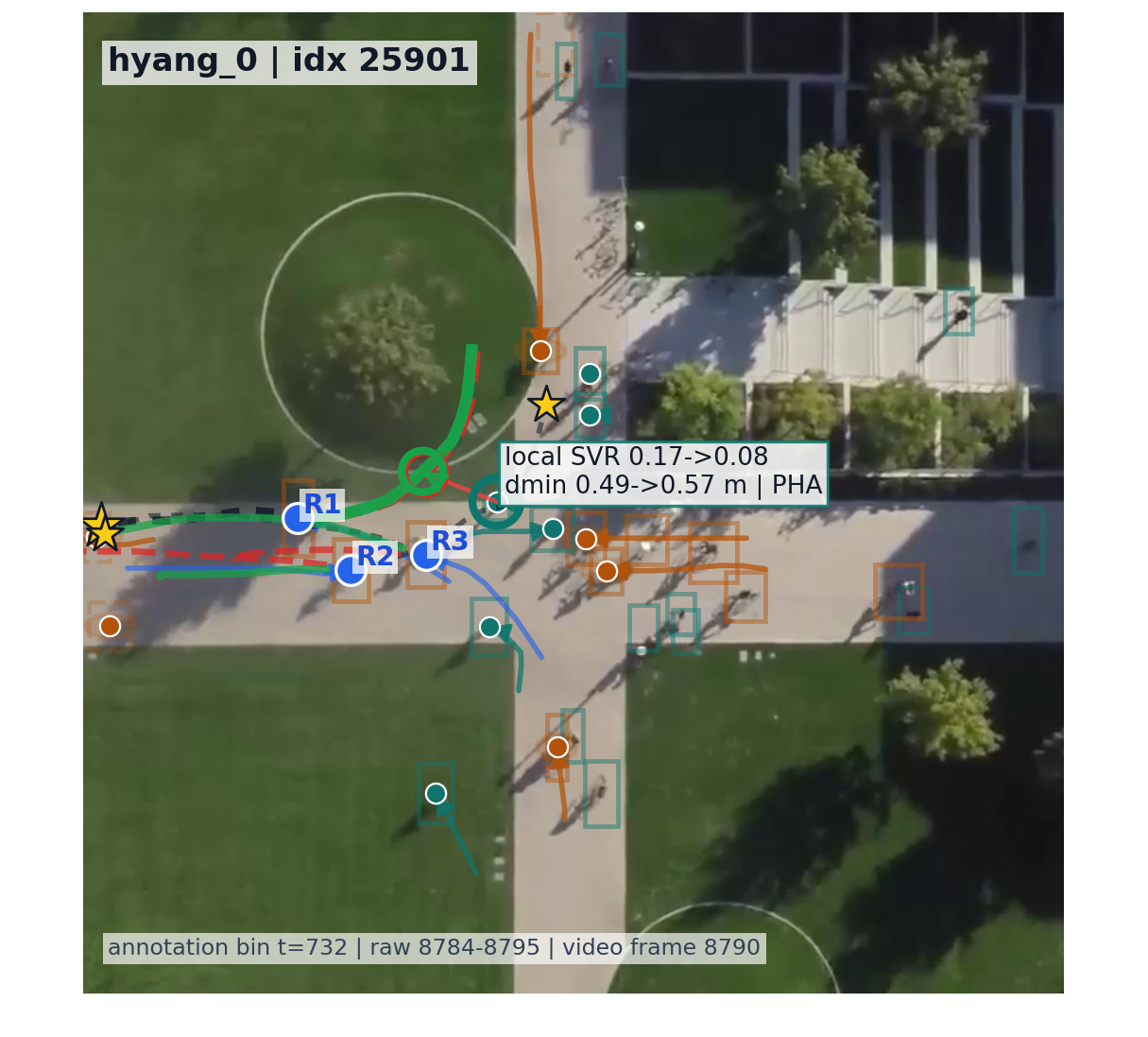}\hspace{0.5mm}%
\raisebox{-0.5mm}{%
\includegraphics[width=0.17\textwidth, trim=17 10 30 5, clip]
{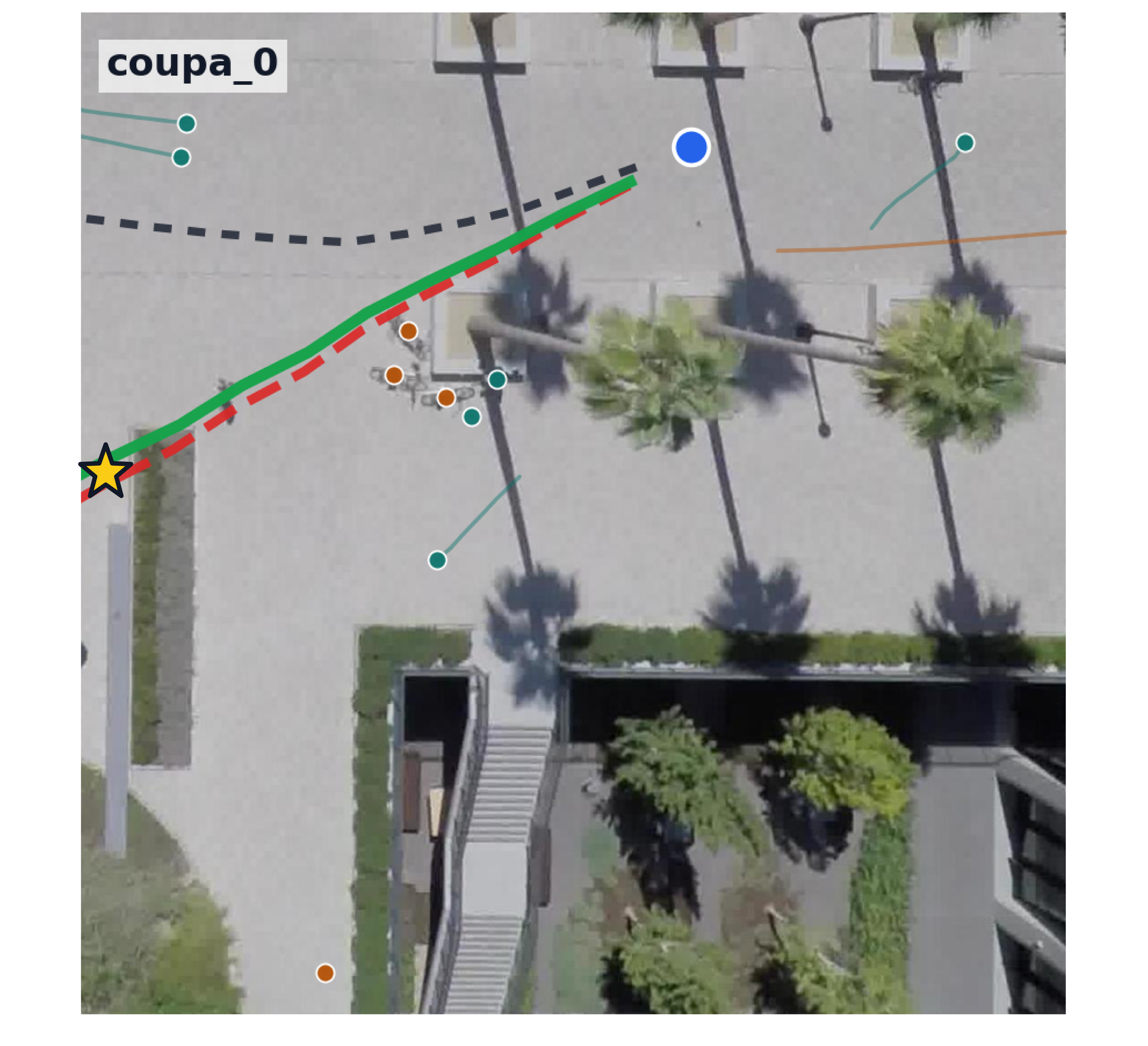}%
}\hspace{0.5mm}%
\includegraphics[width=0.15\textwidth, trim=17 10 30 5, clip]
{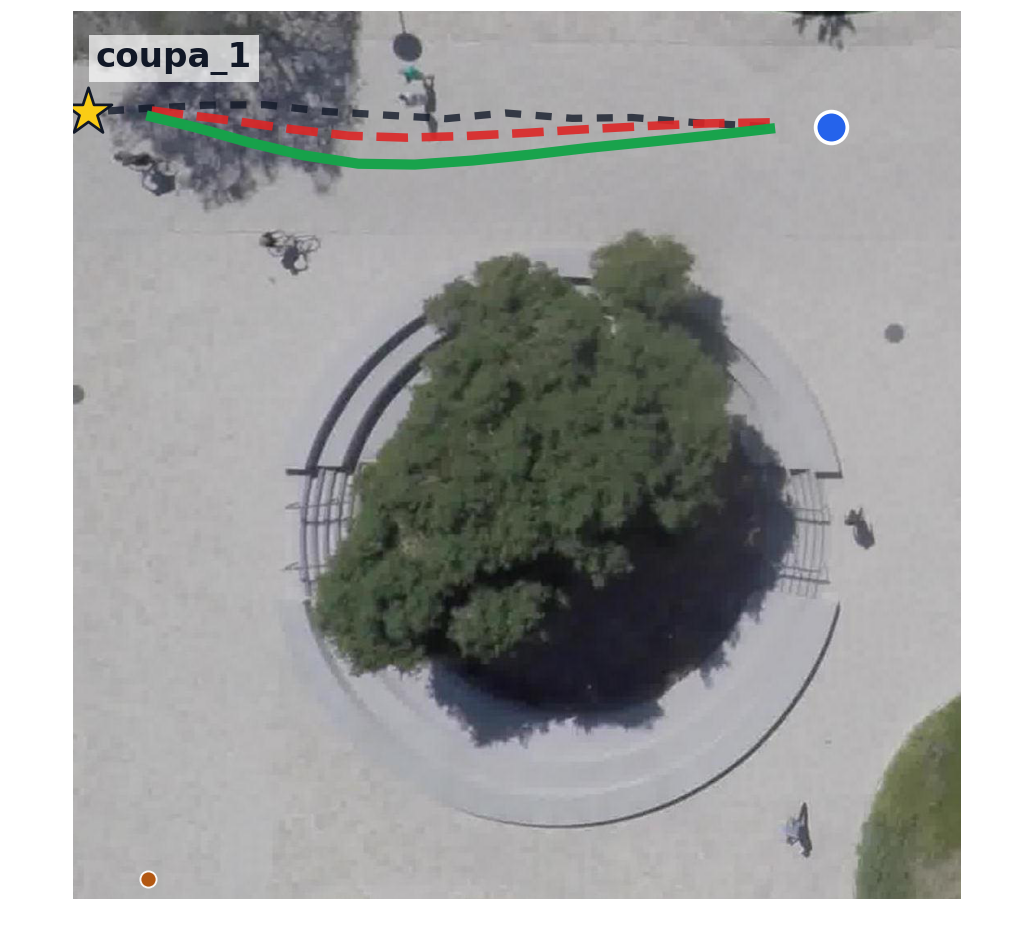}
\vspace{-3mm}
\caption{Qualitative comparisons in SDD scenes, where \texttt{hyang\_0}, \texttt{coupa\_0}, and \texttt{coupa\_1} are official SDD video-scene identifiers. The first picture shows three robots in \texttt{hyang\_0}; the last two pictures show a single robot in \texttt{coupa\_0} and \texttt{coupa\_1}. Blue circles and yellow stars denote robots and waypoints, respectively; black, red, and green lines denote recorded futures, unguided trajectories, and SAGE-guided trajectories, respectively. Colored markers and thin traces show surrounding role-labeled entities.}
\label{fig:sdd_qualitative}
\vspace{-5mm}
\end{figure}

Fig.~\ref{fig:synthetic_qualitative} shows representative multi-robot cases in controlled simulations, where only the guidance setting changes. Three patterns emerge. \textit{First}, guidance primarily adjusts trajectories near social fields or collision boundaries, while leaving other trajectories largely unchanged. \textit{Second}, robots maintain larger clearance from PHAs than from SSAAs, reflecting the role-conditioned social fields. \textit{Third}, guided trajectories remain close to their assigned waypoints, indicating that improved safety and social compliance do not substantially compromise waypoint progress. Fig.~\ref{fig:sdd_qualitative} shows similar behavior on SDD dataset, where guidance redirects robot trajectories away from nearby heterogeneous entities.

\vspace{-2mm}
% \subsection{Additional Diagnostics}

% Appendix~\ref{sec:app_ablations} evaluates heterogeneous encoding, joint versus two-stage generation, task-progress weighting, and pseudo-role robustness. Appendix~\ref{sec:app_baselines} covers deterministic Social Force planning, recorded SDD multi-robot proxies, and scale-and-density diagnostics using locally generated social-navigation layouts, complementing the main evaluation of representation, prediction--planning coupling, task control, robustness, and cross-regime behavior.

\vspace{-3mm}
\subsection{Additional Ablations and Robustness Studies}
\label{sec:additional_results}

To further assess the robustness of SAGE and isolate its key design choices, Appendix~A provides additional ablations and diagnostics in Tables~\ref{tab:ethucy_full}--\ref{tab:socialgym}. In the following, we briefly discuss the key findings and refer interested readers to Appendix~A for further details and discussion. In brief, the per-scene ETH/UCY results in Table~\ref{tab:ethucy_full}, the complete SDD results in Table~\ref{tab:sdd_full}, and the role-wise SDD results in Table~\ref{tab:sdd_rolewise} show that, when comparing SAGE w/ and w/o guidance, guidance consistently improves safety and social compliance across different scenes and semantic categories, while largely preserving trajectory prediction accuracy and task performance. Table~\ref{tab:ablation} further shows that replacing HGT with a homogeneous graph-attention encoder slightly improves displacement accuracy (i.e., R-ADE and Goal-FDE) but worsens CR, SVR, and Energy, confirming the benefit of role-dependent interaction modeling. Moreover, Table~\ref{tab:decoupled} shows that jointly generating robot and entity trajectories achieves better waypoint-directed performance as the robot team grows compared with a two-stage approach that first predicts entity trajectories and then plans robot trajectories based on the fixed predictions, highlighting the benefit of coupling trajectory prediction and robot planning within a unified generative framework.
Additional results demonstrate the controllability and generalizability of SAGE: Table~\ref{tab:wgoal} shows that increasing $w_{\mathrm{goal}}$ improves waypoint-reaching performance at the cost of higher safety and social-compliance violations, revealing an interpretable task-safety trade-off. Table~\ref{tab:role_robustness} confirms that the safety benefits of guided sampling remain consistent under different pseudo-role construction strategies on ETH/UCY. Compared with the Social Force planner~\cite{helbing1995social} in Table~\ref{tab:socialforce}, SAGE achieves substantially better task progress and trajectory accuracy, while Social Force attains lower violation rates through more conservative repulsion. Finally, the multi-robot SDD results in Table~\ref{tab:sdd_multirobot} and Table~\ref{tab:socialgym} demonstrate that SAGE generalizes beyond the primary controlled-simulation setting, with guidance consistently improving robot-entity safety, inter-robot safety, and social compliance.

\vspace{-2.5mm}
\section{Conclusion and Future Work}
In this work, we introduced SAGE for safe and socially compliant robot navigation in heterogeneous multi-agent environments. SAGE models role-dependent and asymmetric agent interactions through a directed heterogeneous graph and an HGT-based scene encoder. Using the resulting interaction-aware representations, a conditional diffusion model jointly generates predicted entity trajectories and planned robot trajectories. SAGE further integrates a training-free safety-social energy guidance mechanism that incorporates collision avoidance, kinematic feasibility, task progress, and role-dependent social norms into the reverse sampling process, enabling robot trajectories to be refined at inference time without retraining. Our evaluations in real-world and controlled heterogeneous multi-robot scenarios demonstrated that SAGE improves physical safety and social compliance while largely preserving trajectory accuracy and task performance. Promising future directions include \textit{(i)} extending SAGE to decentralized multi-robot navigation under partial observability and communication constraints and \textit{(ii)} adapting role-dependent social priors to capture context-, culture-, and individual-dependent preferences through adaptive or human-in-the-loop mechanisms.

% SAGE combines HGT interaction encoding, joint diffusion prediction/planning, and training-free safety-social guidance. Recorded and controlled experiments show improved safety/social compliance with preserved task progress and a controllable trade-off. Future work will target adaptive sampling and physical-robot validation.
\vspace{-4mm}
{\scriptsize

\bibliographystyle{IEEEtran}
\bibliography{ref}
}

\clearpage
\twocolumn
\appendices

\section{Additional Experimental Details and Results}
\label{app:aux_details}
\suppressfloats[t]

This appendix collects implementation settings and additional results. 

% \subsection{Implementation and Guidance Settings}
% \label{sec:supp_settings}

% ETH/UCY retains up to eight neighboring entities and SDD up to 12 in single-robot settings. ETH/UCY and controlled simulation use metric coordinates, whereas SDD coordinates are divided by 50 before training and evaluation; thresholds are expressed in these respective coordinate systems. The tuple $(d^{(\mathsf{safe})},v_{\max},d^{(\mathsf{goal})})$ is $(0.45,2.5,0.5)$ for ETH/UCY, $(0.30,3.0,0.5)$ for controlled simulation, and $(0.35,3.0,0.6)$ for SDD; $v_{\max}^{\mathsf{lat}}=0.15$ and $\omega_{\max}=1.2$ are shared. Following proxemics theory~\cite{hall1966hidden} and social-navigation literature~\cite{helbing1995social}, $(\sigma^{\mathsf{lon}},\sigma^{\mathsf{lat}})$ is $(2.0,1.0)$ for PHAs, $(2.6,1.2)$ for HNAAs, and $(1.2,0.7)$ for SSAAs.

% Unless stated, $(w_{\mathsf{col}},w_{\mathsf{soc}},w_{\mathsf{kin}},w_{\mathsf{goal}})=(2.0,1.0,0.1,0.3)$, $\sigma_{\mathsf{col}}=0.8$, and the gradient-clipping and guidance-step-clipping norms are $1.0$ and $0.05$, respectively; $w_{\mathsf{R2R}}=1.5$ in controlled simulation and $1.0$ in the SDD multi-robot proxy. AdamW uses learning rate $3\times10^{-4}$, weight decay $10^{-4}$, and batch size 128. ETH/UCY and SDD models train for 100 and 80 epochs, respectively. Each controlled-simulation model trains for 50 epochs on 8,192 episodes and is evaluated on 1,024 held-out episodes per seed. The HGT has $L=2$, $d=128$, and four heads; the denoiser has four temporal Transformer layers and 100 diffusion steps.

\subsection{Implementation and Guidance Settings}
\label{sec:supp_settings}

We retain up to eight neighboring entities in ETH/UCY and up to 12 in SDD. ETH/UCY and the controlled simulations use metric coordinates, whereas SDD coordinates are divided by 50 before training and evaluation; accordingly, all distance-related thresholds are specified in their respective coordinate systems. The safety distance, maximum velocity, and goal-reaching threshold $(d^{(\mathsf{safe})},v_{\max},d^{(\mathsf{goal})})$ are set to $(0.45,2.5,0.5)$ for ETH/UCY, $(0.30,3.0,0.5)$ for the controlled simulations, and $(0.35,3.0,0.6)$ for SDD. Across all settings, we use $v_{\max}^{\mathsf{lat}}=0.15$ and $\omega_{\max}=1.2$. Following proxemics theory~\cite{hall1966hidden} and prior social-navigation literature~\cite{helbing1995social}, the role-dependent longitudinal and lateral social margins $(\sigma^{\mathsf{lon}},\sigma^{\mathsf{lat}})$ are set to $(2.0,1.0)$ for PHAs, $(2.6,1.2)$ for HNAAs, and $(1.2,0.7)$ for SSAAs.

Unless otherwise stated, the guidance weights are set to $(w_{\mathsf{col}},w_{\mathsf{soc}},w_{\mathsf{kin}},w_{\mathsf{goal}})=(2.0,1.0,0.1,0.3)$, with $\sigma_{\mathsf{col}}=0.8$. The gradient-clipping and guidance-step-clipping norms are set to $1.0$ and $0.05$, respectively, while the robot-to-robot interaction weight $w_{\mathsf{R2R}}$ is set to $1.5$ in the controlled simulations and $1.0$ in the SDD multi-robot  experiments.

For model training, we use AdamW with a learning rate of $3\times10^{-4}$, a weight decay of $10^{-4}$, and a batch size of 128. The ETH/UCY and SDD models are trained for 100 and 80 epochs, respectively. Each controlled-simulation model is trained for 50 epochs using 8,192 episodes and evaluated on 1,024 held-out episodes per seed. The HGT consists of $L=2$ layers with hidden dimension $d=128$ and four attention heads, while the denoiser consists of four temporal Transformer layers and uses 100 diffusion steps.

% \begin{table}[H]
% \centering
% \caption{Guidance-strength sweep on ETH/UCY (five-scene average).}
% \label{tab:guidance_scale}
% \scriptsize
% \setlength{\tabcolsep}{2.0pt}
% \begin{tabular}{cccccccc}
% \hline
% \rowcolor{gray!25}  $\lambda$ & \textbf{R-ADE}$\downarrow$ & \textbf{Goal-FDE}$\downarrow$ & \textbf{CR}$\downarrow$ & \textbf{PIR}$\downarrow$ & \textbf{SVR}$\downarrow$ & \textbf{Energy}$\downarrow$ \\
% \hline
% 0.00 & 0.1152 & 0.0937 & 0.0441 & 0.3560 & 0.1479 & 20.3761 \\
% 0.03 & 0.1160 & 0.0946 & 0.0390 & 0.3497 & 0.1324 & 19.5015 \\
% 0.05 & 0.1174 & 0.0947 & 0.0353 & 0.3457 & 0.1248 & 19.1257 \\
% \hline
% \end{tabular}
% \end{table}

\subsection{Complete Recorded-Scene Diagnostics}
\label{sec:supp_recorded_diagnostics}

% Table~\ref{tab:ethucy_full} reports the full ETH/UCY per-scene proxy-navigation results underlying the averages in Table~\ref{tab:main_results}. Table~\ref{tab:sdd_full} provides the complete SDD aggregate result including entity displacement, while Table~\ref{tab:sdd_rolewise} gives the role-wise decomposition.

\begin{table*}[!b]
\centering
\caption{Full ETH/UCY per-scene proxy-navigation results.}
\label{tab:ethucy_full}
\scriptsize
\begin{tabular}{llcccccccccc}
\hline
\rowcolor{gray!25}
\textbf{Dataset} & \textbf{Method} & \textbf{R-ADE}$\downarrow$ & \textbf{Goal-FDE}$\downarrow$ & \textbf{Goal-SR}$\uparrow$ & \textbf{E-ADE}$\downarrow$ & \textbf{E-FDE}$\downarrow$ & \textbf{CR}$\downarrow$ & \textbf{MD}$\uparrow$ & \textbf{PIR}$\downarrow$ & \textbf{SVR}$\downarrow$ & \textbf{Energy}$\downarrow$ \\
\hline
ETH & SAGE w/o guidance & 0.1338 & 0.0550 & 1.0000 & 0.7319 & 1.2722 & 0.0723 & 0.0362 & 0.3580 & 0.1494 & 10.5465 \\
ETH & SAGE w/ guidance  & 0.1339 & 0.0531 & 1.0000 & 0.7183 & 1.2350 & 0.0591 & 0.0549 & 0.3538 & 0.1221 & 9.9119 \\
HOTEL & SAGE w/o guidance & 0.0977 & 0.1132 & 0.9691 & 0.5364 & 1.0041 & 0.0491 & 0.0551 & 0.3884 & 0.1000 & 10.0647 \\
HOTEL & SAGE w/ guidance  & 0.0997 & 0.1116 & 0.9783 & 0.5452 & 1.0207 & 0.0392 & 0.0684 & 0.3733 & 0.0802 & 9.2323 \\
UNIV & SAGE w/o guidance & 0.1567 & 0.1441 & 0.9681 & 0.7837 & 1.4014 & 0.0393 & 0.0213 & 0.2574 & 0.1298 & 35.4965 \\
UNIV & SAGE w/ guidance & 0.1588 & 0.1467 & 0.9686 & 0.7835 & 1.4011 & 0.0341 & 0.0258 & 0.2482 & 0.1170 & 33.5777 \\
ZARA1 & SAGE w/o guidance & 0.0954 & 0.0661 & 0.9996 & 0.2899 & 0.4944 & 0.0240 & 0.0707 & 0.4006 & 0.1869 & 16.9963 \\
ZARA1 & SAGE w/ guidance  & 0.0993 & 0.0711 & 1.0000 & 0.2860 & 0.4893 & 0.0179 & 0.0942 & 0.3849 & 0.1722 & 15.9966 \\
ZARA2 & SAGE w/o guidance & 0.0926 & 0.0900 & 0.9995 & 0.3076 & 0.5763 & 0.0358 & 0.0520 & 0.3753 & 0.1733 & 28.7766 \\
ZARA2 & SAGE w/ guidance & 0.0955 & 0.0908 & 0.9993 & 0.3057 & 0.5707 & 0.0265 & 0.0680 & 0.3684 & 0.1326 & 26.9102 \\
\hline
Avg. & SAGE w/o guidance & 0.1152 & 0.0937 & 0.9873 & 0.5299 & 0.9497 & 0.0441 & 0.0471 & 0.3560 & 0.1479 & 20.3761 \\
Avg. & SAGE w/ guidance & 0.1174 & 0.0947 & 0.9892 & 0.5278 & 0.9434 & 0.0353 & 0.0623 & 0.3457 & 0.1248 & 19.1257 \\
\hline
\end{tabular}
\end{table*}

\begin{table*}[!t]
\centering
\caption{Full SDD semantic diagnostics with official annotations ($N_{\mathrm{samp}}=20$ samples).}
\label{tab:sdd_full}
\scriptsize
\setlength{\tabcolsep}{2.0pt}
\begin{tabular}{lccccccccccc}
\hline
\rowcolor{gray!25}
\textbf{Method} & \textbf{R-ADE}$\downarrow$ & \textbf{Goal-FDE}$\downarrow$ & \textbf{Goal-SR}$\uparrow$ & \textbf{E-ADE}$\downarrow$ & \textbf{E-FDE}$\downarrow$ & \textbf{CR}$\downarrow$ & \textbf{MD}$\uparrow$ & \textbf{PIR}$\downarrow$ & \textbf{SVR}$\downarrow$ & \textbf{Sp-VR}$\downarrow$ & \textbf{Energy}$\downarrow$ \\
\hline
SAGE w/o guidance & 0.1151 & 0.0700 & 0.9945 & 0.4769 & 0.9274 & 0.1763 & 0.0582 & 0.1612 & 0.0925 & 0.0298 & 26.1494 \\
SAGE  & 0.1241 & 0.0813 & 0.9913 & 0.4768 & 0.9274 & 0.1729 & 0.0694 & 0.1576 & 0.0679 & 0.0300 & 24.4608 \\
\hline
\end{tabular}
\end{table*}

\begin{table}[t]
\centering
\caption{SDD role-wise semantic diagnostics.}
\label{tab:sdd_rolewise}
\scriptsize
\setlength{\tabcolsep}{1.8pt}
\begin{tabular}{llccccccc}
\hline
\rowcolor{gray!25}
\textbf{Role} & \textbf{Guidance} & \textbf{ADE}$\downarrow$ & \textbf{FDE}$\downarrow$ & \textbf{CR}$\downarrow$ & \textbf{MD}$\uparrow$ & \textbf{PIR}$\downarrow$ & \textbf{SVR}$\downarrow$ & \textbf{Count/Scene} \\
\hline
PHA & w/o & 0.2781 & 0.5223 & 0.0141 & 0.1329 & 0.1786 & 0.0933 & 2.6130 \\
PHA & w/  & 0.2787 & 0.5230 & 0.0119 & 0.1476 & 0.1748 & 0.0710 & 2.6130 \\
HNAA & w/o & 0.7382 & 1.4622 & 0.0134 & 0.1915 & 0.1040 & 0.0752 & 1.7265 \\
HNAA & w/  & 0.7385 & 1.4631 & 0.0123 & 0.2098 & 0.1008 & 0.0508 & 1.7265 \\
SSAA & w/o & 0.1428 & 0.2569 & 0.0058 & 0.6455 & 0.0327 & 0.0103 & 0.0697 \\
SSAA & w/  & 0.1406 & 0.2526 & 0.0057 & 0.6475 & 0.0320 & 0.0042 & 0.0697 \\
\hline
\end{tabular}
\end{table}

% Guidance reduces role-wise CR by 15.6\% for PHAs (0.0141$\rightarrow$0.0119), 8.2\% for HNAAs (0.0134$\rightarrow$0.0123), and marginally for SSAAs. SVR decreases by 23.9\% (PHA, 0.0933$\rightarrow$0.0710), 32.4\% (HNAA, 0.0752$\rightarrow$0.0508), and 59.2\% (SSAA, 0.0103$\rightarrow$0.0042). The improvement is distributed across all semantic categories rather than concentrated in one dominant class. SSAA samples are rare in the evaluated split (0.07 per scene), so their role-wise numbers should be interpreted as diagnostic rather than conclusive.

Table~\ref{tab:ethucy_full} reports the full per-scene ETH/UCY proxy-navigation results underlying the averages in Table~\ref{tab:main_results}. Across all five scenes, guidance consistently reduces CR, PIR, SVR, and Energy while increasing MD, with only minor changes in trajectory accuracy and goal-reaching performance. On average, CR decreases from 0.0441 to 0.0353 and SVR from 0.1479 to 0.1248, while MD increases from 0.0471 to 0.0623. Meanwhile, R-ADE changes only from 0.1152 to 0.1174, and Goal-SR slightly improves from 0.9873 to 0.9892. These per-scene results confirm that the safety and social-compliance gains reported in Table~\ref{tab:main_results} are consistent across the ETH/UCY scenes rather than driven by a particular environment.

Table~\ref{tab:sdd_full} provides the complete aggregate results on SDD, including entity displacement metrics. Guidance reduces CR from 0.1763 to 0.1729, PIR from 0.1612 to 0.1576, SVR from 0.0925 to 0.0679, and Energy from 26.1494 to 24.4608, while increasing MD from 0.0582 to 0.0694. Notably, entity trajectory prediction remains essentially unchanged, with E-ADE changing from 0.4769 to 0.4768 and E-FDE remaining at 0.9274. These safety and social-compliance improvements come with a modest degradation in robot trajectory and goal-reaching metrics, as R-ADE increases from 0.1151 to 0.1241, Goal-FDE from 0.0700 to 0.0813, and Goal-SR decreases slightly from 0.9945 to 0.9913.

Finally, Table~\ref{tab:sdd_rolewise} decomposes the SDD results across the three semantic agent categories. Guidance reduces role-wise CR by 15.6\% for PHAs (0.0141$\rightarrow$0.0119) and 8.2\% for HNAAs (0.0134$\rightarrow$0.0123), with a marginal reduction for SSAAs (0.0058$\rightarrow$0.0057). More pronounced improvements are observed in social compliance: SVR decreases by 23.9\% for PHAs (0.0933$\rightarrow$0.0710), 32.4\% for HNAAs (0.0752$\rightarrow$0.0508), and 59.2\% for SSAAs (0.0103$\rightarrow$0.0042). Also, upon using guidance, MD  increases and PIR decreases for all three categories, while their ADE and FDE remain nearly unchanged. Thus, the benefits of guidance extend across all semantic categories rather than being concentrated in a single dominant class. Since SSAAs are relatively rare in the evaluated split (0.07 per scene), their role-wise results should be interpreted as diagnostic rather than conclusive.

\begin{figure}[!t]
    \centering
    \includegraphics[
        width=\columnwidth,
        pagebox=crop
    ]{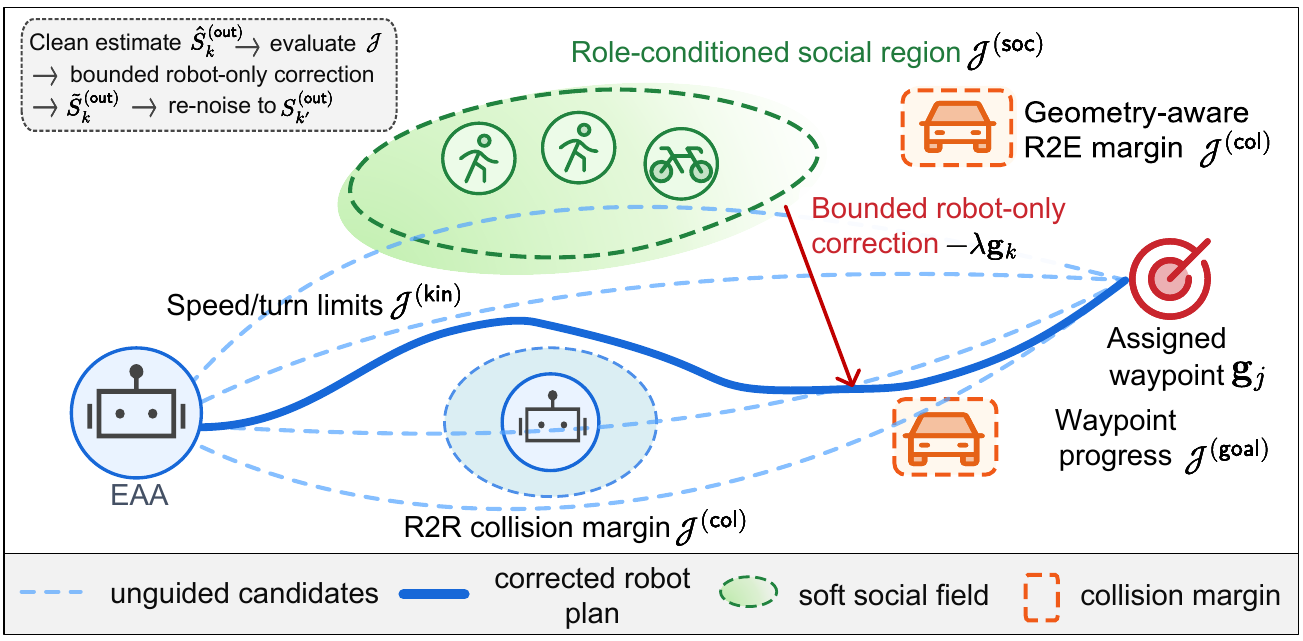}
    \caption{Detailed illustration of one safety-social guidance step
    during reverse diffusion. The role-conditioned social field,
    geometry-aware R2E/R2R collision margins, kinematic limits, and
    waypoint progress jointly define the guidance energy. Its clipped
    gradient applies a bounded correction only to the robot trajectory
    before re-noising, while entity predictions remain unchanged.}
    \label{fig:app_guidance}
\end{figure}

\subsection{Additional Ablations and Diagnostics}
\label{sec:app_ablations}

\subsubsection{Heterogeneous encoder}
To assess the contribution of heterogeneous interaction modeling, we replace the HGT with a homogeneous graph-attention encoder that uses a single shared relation type and omits role embeddings, thereby removing relation-specific attention projections. As shown in Table~\ref{tab:ablation}, the homogeneous encoder achieves better displacement accuracy, with R-ADE and Goal-FDE decreasing from 0.1241 to 0.1170 and from 0.0813 to 0.0621, respectively. In contrast, HGT provides better safety and social-compliance performance, reducing CR from 0.1806 to 0.1729, SVR from 0.0743 to 0.0679, and Energy from 24.8181 to 24.4608. These results indicate that explicitly modeling role-dependent and relation-specific interactions primarily improves interaction quality and social compliance, albeit with a modest trade-off in raw trajectory accuracy.

\begin{table}[b!]
\centering
\caption{HGT versus homogeneous graph-attention encoding on SDD under guided sampling.}
\label{tab:ablation}
\scriptsize
\setlength{\tabcolsep}{1.5pt}
\begin{tabular}{lccccc}
\hline
\rowcolor{gray!25}
\textbf{Variant} & \textbf{R-ADE}$\downarrow$ & \textbf{Goal-FDE}$\downarrow$ & \textbf{CR}$\downarrow$ & \textbf{SVR}$\downarrow$ & \textbf{Energy}$\downarrow$ \\
\hline
Homogeneous encoder & 0.1170 & 0.0621 & 0.1806 & 0.0743 & 24.8181 \\
HGT & 0.1241 & 0.0813 & 0.1729 & 0.0679 & 24.4608 \\
\hline
\end{tabular}
\end{table}

\subsubsection{Joint vs.\ two-stage generation}
To evaluate the benefit of jointly generating robot and surrounding-entity trajectories, Table~\ref{tab:decoupled} compares SAGE with a two-stage variant in which entity trajectories are first generated and then fixed while robot trajectories are subsequently generated using the same Phase~3 guidance settings. The two-stage variant achieves slightly better local safety and social-compliance metrics, with lower R2E-CR, R2R-CR, and SVR and notably higher R2R-MD. In contrast, joint generation consistently achieves better Goal-FDE, reducing it from 5.725 to 5.448 for $N_r=12$ and from 5.675 to 5.444 for $N_r=20$. These results suggest that, while decoupling prediction and planning can favor local collision avoidance, jointly modeling the evolution of surrounding entities and robot actions better preserves goal-directed task performance as multiple robots plan simultaneously.

\begin{table}[t]
\centering
\caption{Multi-robot two-stage decoupling in controlled simulation.}
\label{tab:decoupled}
\scriptsize
\setlength{\tabcolsep}{1.5pt}
\resizebox{\columnwidth}{!}{
\begin{tabular}{lllcccccc}
\hline
\rowcolor{gray!25}
\textbf{Setting} & $N_r$ & \textbf{Variant} & \textbf{Guidance} & \textbf{Goal-FDE}$\downarrow$ & \textbf{R2E-CR}$\downarrow$ & \textbf{R2R-CR}$\downarrow$ & \textbf{R2R-MD}$\uparrow$ & \textbf{SVR}$\downarrow$ \\
\hline
Controlled sim. & 12 & Joint (i.e., SAGE) & w/  & \textbf{5.448} & 0.0379 & 0.0839 & 0.0207 & 0.0191 \\
Controlled sim. & 12 & Two-stage & w/  & 5.725 & \textbf{0.0369} & \textbf{0.0788} & \textbf{0.0626} & \textbf{0.0178} \\
Controlled sim. & 20 & Joint (i.e., SAGE) & w/  & \textbf{5.444} & 0.0368 & 0.0769 & 0.0141 & 0.0183 \\
Controlled sim. & 20 & Two-stage & w/  & 5.675 & \textbf{0.0360} & \textbf{0.0757} & \textbf{0.0391} & \textbf{0.0173} \\
\hline
\end{tabular}
}
\end{table}

% The two-stage variant is competitive on local safety metrics, particularly R2R-MD where it substantially exceeds joint generation. However, the joint model preserves better Goal-FDE in large-team settings (e.g., 5.444 vs.\ 5.675 at $N_r=20$), indicating that coupled generation helps maintain task consistency when multiple robots plan simultaneously.

\subsubsection{Task-progress recovery via $w_{\mathsf{goal}}$}
To characterize the trade-off between goal-directed task performance and safety/social compliance, Table~\ref{tab:wgoal} sweeps the task-progress weight $w_{\mathsf{goal}}$ on SDD under guided sampling. As $w_{\mathsf{goal}}$ increases from $0$ to $2.0$, R-ADE decreases from 1.4740 to 0.8267, Goal-FDE decreases from 2.4349 to 1.1677, and Goal-SR increases from 0.2120 to 0.3169, indicating progressively stronger goal-directed behavior. This improvement comes at the cost of higher CR (0.0118$\rightarrow$0.0146), PIR (0.0087$\rightarrow$0.0106), SVR (0.0025$\rightarrow$0.0044), and Energy (5.5359$\rightarrow$23.5397). Thus, $w_{\mathsf{goal}}$ provides an explicit control knob for balancing goal-directed task performance against safety and social compliance. Note that this table reports results in raw coordinate units; therefore, the relative trends, rather than direct comparisons with normalized results reported elsewhere, are of primary interest.

\begin{table}[t]
\centering
\caption{SDD $w_{\mathsf{goal}}$ sweep under guided sampling.}
\label{tab:wgoal}
\scriptsize
\setlength{\tabcolsep}{1.8pt}
\begin{tabular}{ccccccccc}
\hline
\rowcolor{gray!25}
$w_{\mathsf{goal}}$ & \textbf{R-ADE}$\downarrow$ & \textbf{Goal-FDE}$\downarrow$ & \textbf{Goal-SR}$\uparrow$ & \textbf{CR}$\downarrow$ & \textbf{PIR}$\downarrow$ & \textbf{SVR}$\downarrow$ & \textbf{Energy}$\downarrow$ \\
\hline
0.0 & 1.4740 & 2.4349 & 0.2120 & 0.0118 & 0.0087 & 0.0025 & 5.5359 \\
0.3 & 1.3487 & 2.1951 & 0.2247 & 0.0122 & 0.0090 & 0.0027 & 11.4233 \\
0.6 & 1.2335 & 1.9741 & 0.2387 & 0.0126 & 0.0092 & 0.0029 & 15.9795 \\
1.0 & 1.0945 & 1.7031 & 0.2595 & 0.0131 & 0.0096 & 0.0033 & 19.8706 \\
2.0 & 0.8267 & 1.1677 & 0.3169 & 0.0146 & 0.0106 & 0.0044 & 23.5397 \\
\hline
\end{tabular}
\end{table}

% Increasing $w_{\mathsf{goal}}$ from 0 to 2.0 reduces Goal-FDE from 2.4349 to 1.1677 and improves Goal-SR from 0.2120 to 0.3169, at the cost of gradually higher CR, PIR, SVR, and Energy. This confirms $w_{\mathsf{goal}}$ as a tunable task--safety control knob.

% \textit{Pseudo-role construction robustness.}
% ETH/UCY lacks semantic labels. Table~\ref{tab:role_robustness} verifies that the guidance effect does not depend on a particular pseudo-labeling rule. Here, \texttt{speed\_threshold} and \texttt{speed\_quantile} assign roles using fixed thresholds and speed quantiles, respectively, whereas \texttt{all\_pha} labels every entity as PHA.

\subsubsection{Pseudo-role construction robustness}
Since ETH/UCY does not provide semantic agent labels, Table~\ref{tab:role_robustness} evaluates whether the effectiveness of guidance depends on the pseudo-role construction strategy. We consider three alternatives: \texttt{speed\_threshold}, which assigns roles using fixed speed thresholds; \texttt{speed\_quantile}, which uses speed quantiles; and \texttt{all\_pha}, which assigns every entity to the PHA category. Across all three strategies, guidance consistently and substantially reduces CR, PIR, SVR, and Energy. For example, CR decreases from approximately $0.046$--$0.047$ without guidance to $0.0030$--$0.0034$ with guidance, while SVR decreases from $0.1673$--$0.2083$ to $0.0218$--$0.0261$. Similarly, PIR decreases from $0.3625$--$0.3885$ to $0.0961$--$0.1018$, and Energy from $25.7617$--$27.5432$ to $9.4370$--$9.6635$. These consistent improvements demonstrate that the effectiveness of guidance is robust to the particular pseudo-role construction strategy used for ETH/UCY. Since these experiments use separate task-conditioned checkpoints with $\lambda=0.03$, only guided-versus-unguided comparisons within each strategy should be made; their absolute values are not directly comparable with those in Table~\ref{tab:main_results}.

\begin{table}[H]
\centering
\caption{Proxy role construction ablation on ETH/UCY (five-scene average, $N_{\mathrm{samp}}=20$ samples).}
\label{tab:role_robustness}
\scriptsize
\setlength{\tabcolsep}{1.5pt}
\begin{tabular}{llcccccc}
\hline
\rowcolor{gray!25}
\textbf{Role Strategy} & \textbf{Guidance} & \textbf{R-ADE}$\downarrow$ & \textbf{Goal-FDE}$\downarrow$ & \textbf{CR}$\downarrow$ & \textbf{PIR}$\downarrow$ & \textbf{SVR}$\downarrow$ & \textbf{Energy}$\downarrow$ \\
\hline
\texttt{speed\_threshold} & w/o & 0.5149 & 0.8900 & 0.0456 & 0.3885 & 0.1673 & 26.4692 \\
\texttt{speed\_threshold} & w/  & 0.6980 & 1.0756 & 0.0030 & 0.1018 & 0.0218 & 9.4370 \\
\texttt{speed\_quantile}  & w/o & 0.6054 & 1.0353 & 0.0459 & 0.3625 & 0.1714 & 25.7617 \\
\texttt{speed\_quantile}  & w/  & 0.6748 & 1.0259 & 0.0034 & 0.0961 & 0.0261 & 9.6635 \\
\texttt{all\_pha}         & w/o & 0.5519 & 0.9574 & 0.0471 & 0.3810 & 0.2083 & 27.5432 \\
\texttt{all\_pha}         & w/  & 0.7024 & 1.0839 & 0.0031 & 0.0989 & 0.0252 & 9.6558 \\
\hline
\end{tabular}
\end{table}

% Across all strategies, guided sampling sharply reduces CR, PIR, SVR, and Energy. The safety improvement is robust to the choice of pseudo-labeling rule. These separate task-conditioned checkpoints use $\lambda=0.03$; only within-strategy guided/unguided changes, not absolute values against Table~\ref{tab:main_results}, are comparable.

\subsection{Social Force Baseline and Multi-Robot Diagnostics}
\label{sec:app_baselines}

\subsubsection{Social Force comparison}
Table~\ref{tab:socialforce} compares SAGE with the deterministic Social Force planner~\cite{helbing1995social}, which uses hand-crafted attractive and repulsive potentials to regulate robot motion. In the controlled simulation, Social Force achieves lower R2E-CR (0.0136 vs.\ 0.0371), R2R-CR (0.0172 vs.\ 0.0772), and SVR (0.0049 vs.\ 0.0185) than SAGE, but at a substantial cost in goal-directed and trajectory performance: its Goal-FDE is $1.68\times$ higher (9.371 vs.\ 5.583) and R-ADE is $3.33\times$ higher (2.688 vs.\ 0.808). A similar trend is observed in the SocialGym-style setting\footnote{Here, \textit{SocialGym-style} refers to locally generated doorway, hallway, intersection, roundabout, and open-crowd layouts designed to represent diverse social-navigation scenarios; these experiments do not use rollouts from the SocialGym simulator. \label{foot:Social}
}, where Social Force again achieves lower collision and social-violation rates, while SAGE reduces Goal-FDE from 7.766 to 6.605 and R-ADE from 2.322 to 0.696. Notably, SAGE also achieves zero Sp-VR in both settings, compared with 0.0474 and 0.0366 for Social Force. These results highlight the task--safety trade-off between conservative hand-crafted repulsion and SAGE's learned, task-conditioned trajectory generation.

% \noindent
% \textit{Social Force comparison.}
% Table~\ref{tab:socialforce} contrasts deterministic repulsion with SAGE's learned task-conditioned generation.

\begin{table}[H]
\centering
\caption{Comparison with deterministic Social Force planner.}
\label{tab:socialforce}
\scriptsize
\setlength{\tabcolsep}{1.5pt}
\resizebox{\columnwidth}{!}{
\begin{tabular}{llcccccc}
\hline
\rowcolor{gray!25}
\textbf{Setting} & \textbf{Method} & \textbf{Goal-FDE}$\downarrow$ & \textbf{R-ADE}$\downarrow$ & \textbf{R2E-CR}$\downarrow$ & \textbf{R2R-CR}$\downarrow$ & \textbf{SVR}$\downarrow$ & \textbf{Sp-VR}$\downarrow$ \\
\hline
Controlled sim. & Social Force & 9.371 & 2.688 & \textbf{0.0136} & \textbf{0.0172} & \textbf{0.0049} & 0.0474 \\
Controlled sim. & SAGE & \textbf{5.583} & \textbf{0.808} & 0.0371 & 0.0772 & 0.0185 & \textbf{0.0000} \\
SocialGym-style & Social Force & 7.766 & 2.322 & \textbf{0.0083} & \textbf{0.0198} & \textbf{0.0043} & 0.0366 \\
SocialGym-style & SAGE & \textbf{6.605} & \textbf{0.696} & 0.0292 & 0.0823 & 0.0168 & \textbf{0.0000} \\
\hline
\end{tabular}
}
\end{table}

% The Social Force planner~\cite{helbing1995social} aggressively repels robots from surrounding participants using hand-crafted potentials. It achieves low collision and social-violation rates but at a severe task cost: across the controlled-simulation sweep, its Goal-FDE is 1.68$\times$ worse and Robot ADE is 3.33$\times$ worse than SAGE. This diagnostic exposes the task--safety trade-off directly.

% \noindent
% \textit{SDD multi-robot proxy.}
% Table~\ref{tab:sdd_multirobot} reports guided SDD multi-robot proxy results with official semantic annotations. Controlled simulation (Sec.~\ref{subsec:scalability}) provides the primary scalability evidence; these results show that the same pipeline can operate on real semantic scenes.

\subsubsection{SDD multi-robot setting}
Table~\ref{tab:sdd_multirobot} evaluates the SAGE pipeline in multi-robot navigation settings constructed from SDD scenes using the official semantic annotations. As the number of robots increases from $N_r=3$ to $12$, Goal-FDE increases from 0.1260 to 0.3254 and Goal-SR decreases from 0.9716 to 0.8919, reflecting the increasing difficulty of goal-directed navigation with larger robot teams. Notably, the normalized interaction metrics improve with team size (because these metrics are normalized over the increasing number of robot-entity or robot-robot interactions, rather than reporting the total number of violations): R2E-CR decreases from 0.0639 to 0.0130, R2R-CR from 0.2547 to 0.0590, PIR from 0.0567 to 0.0105, and SVR from 0.0296 to 0.0062. Meanwhile, the aggregate Energy increases from 71.46 to 224.74 as more robots contribute to the overall interaction energy. While the controlled simulations in Sec.~\ref{subsec:scalability} provide the primary evidence for scalability, these results demonstrate that the same SAGE pipeline can be extended to multi-robot navigation in real-world scenes with semantic agent annotations.

\begin{table}[H]
\centering
\caption{Guided SDD multi-robot proxy sweep with official semantic annotations.}
\label{tab:sdd_multirobot}
\scriptsize
\setlength{\tabcolsep}{1.5pt}
\begin{tabular}{cccccccc}
\hline
\rowcolor{gray!25}
$N_r$ & \textbf{Goal-FDE}$\downarrow$ & \textbf{Goal-SR}$\uparrow$ & \textbf{R2E-CR}$\downarrow$ & \textbf{R2R-CR}$\downarrow$ & \textbf{PIR}$\downarrow$ & \textbf{SVR}$\downarrow$ & \textbf{Energy}$\downarrow$ \\
\hline
3 & 0.1260 & 0.9716 & 0.0639 & 0.2547 & 0.0567 & 0.0296 & 71.46 \\
6 & 0.2229 & 0.9338 & 0.0249 & 0.1228 & 0.0215 & 0.0128 & 131.63 \\
12 & 0.3254 & 0.8919 & 0.0130 & 0.0590 & 0.0105 & 0.0062 & 224.74 \\
\hline
\end{tabular}
\end{table}

% \noindent
% \textit{SocialGym-style scale and density.}
% Here, SocialGym-style denotes layouts from our local doorway, hallway, intersection, roundabout, and open-crowd generator, not rollouts from the SocialGym simulator. The corresponding SAGE model is trained for 60 epochs on 8,192 generated episodes and evaluated on 1,024 held-out episodes for each of three seeds. Table~\ref{tab:socialgym} reports the resulting scale and density diagnostics.

% Across all SocialGym-style configurations, guidance consistently lowers R2E-CR, R2R-CR, SVR, and Energy, matching the pattern observed in the controlled-simulation scalability sweep.

\subsubsection{SocialGym-style scale and density}
To further stress-test SAGE beyond the primary controlled-simulation setting, we evaluate whether the benefits of guidance persist across a broader range of structured and crowded navigation scenarios and under varying robot-team sizes and entity densities. To this end, we construct \textit{SocialGym-style} environments using our local generator, which produces doorway, hallway, intersection, roundabout, and open-crowd layouts representative of diverse social-navigation scenarios; these experiments do not use rollouts from the SocialGym simulator (see Footnote~\ref{foot:Social}). The corresponding SAGE model is trained for 60 epochs on 8,192 generated episodes and evaluated on 1,024 held-out episodes for each of three seeds. Table~\ref{tab:socialgym} evaluates the effect of guidance across different robot-team sizes and environment densities. Across all configurations, guidance consistently reduces R2E-CR, R2R-CR, SVR, and Energy, with only modest increases in Goal-FDE. For example, at $N_r=6$ and $N_e=80$, guidance reduces R2E-CR from 0.0319 to 0.0296, R2R-CR from 0.0970 to 0.0872, SVR from 0.0194 to 0.0176, and Energy from 789.5 to 740.0, while Goal-FDE increases only from 6.462 to 6.533. Similar trends are observed as the robot team grows from $N_r=3$ to $9$ with $N_e=40$, demonstrating that the safety and social-compliance benefits of guidance persist across different environment structures, team sizes, and entity densities.
\begin{table}[H]
\centering
\caption{SocialGym-style scale and density diagnostics (three evaluation seeds).}
\label{tab:socialgym}
\scriptsize
\setlength{\tabcolsep}{1.5pt}
\begin{tabular}{ccclcccc}
\hline
\rowcolor{gray!25}
$N_r$ & $N_e$ & \textbf{Guidance} & \textbf{Goal-FDE}$\downarrow$ & \textbf{R2E-CR}$\downarrow$ & \textbf{R2R-CR}$\downarrow$ & \textbf{SVR}$\downarrow$ & \textbf{Energy}$\downarrow$ \\
\hline
3 & 40 & w/o & 6.675 & 0.0315 & 0.1050 & 0.0188 & 248.4 \\
3 & 40 & w/  & 6.707 & 0.0292 & 0.0922 & 0.0169 & 233.7 \\
6 & 40 & w/o & 6.543 & 0.0313 & 0.0907 & 0.0185 & 529.6 \\
6 & 40 & w/  & 6.573 & 0.0291 & 0.0807 & 0.0168 & 497.3 \\
6 & 80 & w/o & 6.462 & 0.0319 & 0.0970 & 0.0194 & 789.5 \\
6 & 80 & w/  & 6.533 & 0.0296 & 0.0872 & 0.0176 & 740.0 \\
9 & 40 & w/o & 6.518 & 0.0308 & 0.0853 & 0.0184 & 864.6 \\
9 & 40 & w/  & 6.548 & 0.0288 & 0.0765 & 0.0168 & 815.2 \\
\hline
\end{tabular}
\end{table}
\end{document}